\documentclass[10pt,twocolumn,letterpaper]{article}

\usepackage{wacv}
\usepackage{times}
\usepackage{epsfig}
\usepackage{graphicx}
\usepackage{amsmath}
\usepackage{amssymb}
\usepackage{subcaption}

\usepackage{multirow}
\usepackage[table,xcdraw]{xcolor}

\def\wacvPaperID{****\vspace{-4mm}} 

\wacvfinalcopy 

\ifwacvfinal
\def\assignedStartPage{1} 
\fi

\ifwacvfinal
\usepackage[breaklinks=true,bookmarks=false]{hyperref}
\else
\usepackage[pagebackref=true,breaklinks=true,colorlinks,bookmarks=false]{hyperref}
\fi

\ifwacvfinal
\else
\fi

\begin{document}

\title{PocketNet: Extreme Lightweight Face Recognition Network using Neural Architecture Search and  Multi-Step Knowledge Distillation \vspace{-5mm}}

\author{Fadi Boutros$^{1,2}$, 
Patrick Siebke $^{1}$, 
Marcel Klemt $^{1}$, \\
 Naser Damer$^{1,2}$,
Florian Kirchbuchner$^{1}$, Arjan Kuijper$^{1,2}$\\
$^{1}$Fraunhofer Institute for Computer Graphics Research IGD,
Darmstadt, Germany\\
$^{2}$Mathematical and Applied Visual Computing, TU Darmstadt,
Darmstadt, Germany\\
 Email: {fadi.boutros@igd.fraunhofer.de}
}

\maketitle

\begin{abstract}
Deep neural networks have rapidly become the mainstream method for face recognition (FR). 
However, this limits the deployment of such models that contain an extremely large number of parameters to embedded and low-end devices. 
In this work, we present an extremely lightweight and accurate FR solution, namely PocketNet. 
We utilize neural architecture search to develop a new family of lightweight face-specific architectures. 
We additionally propose a novel training paradigm based on knowledge distillation (KD), the multi-step KD, where the knowledge is distilled from the teacher model to the student model at different stages of the training maturity.
We conduct a detailed ablation study proving both, the sanity of using NAS for the specific task of FR rather than general object classification, and the benefits of our proposed multi-step KD.
We present an extensive experimental evaluation and comparisons with the state-of-the-art (SOTA) compact FR models on nine different benchmarks including large-scale evaluation benchmarks such as IJB-B, IJB-C, and MegaFace.
PocketNets have consistently advanced the SOTA FR performance on nine mainstream benchmarks when considering the same level of model compactness. 
With 0.92M parameters, our smallest network PocketNetS-128 achieved very competitive results to recent SOTA compacted models that contain up to 4M parameters.
Training codes and pre-trained models are made public. \footnote{\url{https://github.com/fdbtrs/PocketNet}}
\end{abstract}

\vspace{-3mm}
\section{Introduction} 
Face recognition is an active research field, and it has benefited from the recent advancements in machine learning, especially the advancements in deep learning \cite{DBLP:conf/cvpr/HeZRS16} and the novelty of margin-based Softmax losses \cite{deng2019arcface,DBLP:conf/cvpr/WangWZJGZL018}, achieving a notable recognition accuracy.
SOTA FR solutions rely on a deep learning models with an extremely large number of parameters \cite{deng2019arcface,MagFace}. 
Deploying such models on embedded devices or in applications with limited memory specifications is a major challenge  \cite{martinez2021benchmarking,DBLP:conf/iccvw/DengGZDLS19}. 
This challenge has received increased attention in the literature in the last few years \cite{martinez2021benchmarking,DBLP:conf/iccvw/DengGZDLS19}.

Recently, several compact FR models have been proposed in the literature.
MobileFaceNet \cite{DBLP:conf/ccbr/ChenLGH18} proposed an efficient FR model based  on MobileNetV2 \cite{DBLP:conf/cvpr/SandlerHZZC18} with around 1M parameters. 
ShuffleFaceNet \cite{DBLP:conf/iccvw/Martinez-DiazLV19} and VarGFaceNet \cite{DBLP:conf/iccvw/YanZXZWS19} model architectures adopted ShuffleNetV2 \cite{DBLP:conf/eccv/MaZZS18} and VarGNet \cite{DBLP:journals/corr/abs-1907-05653}, respectively, for the FR task.
VarGFaceNet contains 5M parameters. ShuffleFaceNet presented three architectures with different width scales (0.5, 1.5 and 2) containing 0.5, 2.6, and 4.5M parameters, respectively. 
MixFaceNets \cite{DBLP:conf/icb/BoutrosDFKK21} use MixNets \cite{DBLP:conf/bmvc/TanL19} as a baseline network structure to develop a new family of FR models. 
The smallest MixFaceNet architecture contains 1.04M parameters and the largest one contains 3.95M parameters. 
Martinez‑Diaz et al. \cite{martinez2021benchmarking} evaluated the computational requirements and the verification performance of five compact model architectures including MobileFaceNet (2.0M parameters), VarGFaceNet \cite{DBLP:conf/iccvw/YanZXZWS19} (5M parameters), ShufeFaceNet \cite{DBLP:conf/iccvw/Martinez-DiazLV19} (2.6m parameters), MobileFaceNetV1 (3.4m parameters), and ProxylessFaceNAS (3.2m parameters). 
The reported results by Martinez‑Diaz et al. \cite{martinez2021benchmarking} demonstrated that compact FR models can still achieve high accuracies for FR.

However, none of these works \cite{DBLP:conf/icb/BoutrosDFKK21,DBLP:conf/ccbr/ChenLGH18,DBLP:conf/iccvw/YanZXZWS19,DBLP:conf/iccvw/Martinez-DiazLV19,martinez2021benchmarking} designed a network specifically for the FR task, rather than adopting existing architectures designed for common computer vision tasks. 
With the developments in AutoML, Neural Architecture Search (NAS) has shown SOTA performances in many computer vision tasks \cite{DBLP:conf/iclr/LiuSY19,DBLP:conf/cvpr/XuG0000O021}. 
NAS aims at automating the neural network architecture design achieving higher performances than the handcraft-designed network architectures.
One of the early works of NAS was introduced by Zoph \etal \cite{zoph2017neural}. That work \cite{zoph2017neural} proposed that the architecture of a neural network can be described as a variable-length string. 
Thus, a Recurrent neural network (RNN) can be used as a controller to generate such a string.
While this method showed competitive results in comparison to SOTA models, it requires a very long search time (22,400 GPU days \cite{zoph2017neural}).
NASNet \cite{zoph2018learning} points out that a convolutional neural network (CNN) such as ResNet \cite{DBLP:conf/cvpr/HeZRS16} is a repetition of modules that consist of combinations of convolution operations. 
Based on that, they introduced a new search space, called NASNet.
NASNet proposed to learn the network building block (cell), rather than learning the whole architecture. The network architecture, in this case, is constructed by stacking these cells together. 
NASNet was able to reduce the search time to 2,000 GPU days, in comparison to 22,400 GPU days needed by NAS \cite{zoph2017neural}.
ProxylessNAS \cite{cai2019proxylessnas} directly learned the architectures for the target task. 
It trained an over-parameterized network by gradient optimization that contained all candidate paths and pruned redundant paths to achieve a compact architecture. The architecture search needed 200 GPU hours, which was much faster than NASNet \cite{zoph2018learning}.
Differentiable Architecture Search (DARTS) \cite{DBLP:conf/iclr/LiuSY19} relaxed the discrete search space in a continuous manner. 
DARTS proposed to use gradient optimization to optimize the architecture search space. 
Similar to NASNet,  DARTS proposed to learn the main building block (cell) of the network architecture rather than learning the entire network architecture.
DARTS search algorithm requires around 1.5 GPU days, which was orders of magnitude faster than NASNet \cite{zoph2018learning}. 
To reduce the NAS search time, all mentioned NAS algorithms proposed to learn from small training datasets such as CIFAR-10 \cite{Krizhevsky09learningmultiple} and then utilized the discovered architecture to train on larger datasets such as ImageNet \cite{DBLP:conf/cvpr/DengDSLL009}.
This advancement in NAS solutions has only recently captured the attention of biometric recognition solutions \cite{DBLP:journals/access/ZhuYK20,DBLP:conf/aaai/Wang21}, however, with no deployments towards lightweight or embedded architectures.

In this work, we successfully aim at intelligently designing and training a family of lightweight FR models, namely the PocketNets, that offer the SOTA trade-off between model compactness and performance. 
To achieve that, we focus on two aspects, the first is the use of a NAS algorithm to learn an FR-specific lightweight network architecture, and the second is to design a novel knowledge distillation (KD) paradigm to relax training difficulties raised by the substantial discrepancy between teacher and student models.
We use CASIA-WebFace (500K images) \cite{DBLP:journals/corr/YiLLL14a} to learn the optimal architecture using DARTS \cite{DBLP:conf/iclr/LiuSY19}.
We additionally propose a novel training paradigm based on KD, namely multi-step KD, to enable transferring the knowledge of the teacher network at different stages of the training process, and thus enhance the verification performance of the compact student model.
We prove our face-specific NAS-based architecture and the proposed multi-step KD in two detailed ablation studies. 
First, we experimentally evaluate the impact of the NAS training dataset source (face vs. general image classes) on the FR performance of the learned architecture.
Second, we experimentally proved and analyzed the competence of our proposed multi-step KD on improving FR performance in comparison to the baseline KD solutions, as well as training without KD.
To experimentally demonstrate the competence of our proposed PocketNets, we report their FR performance on nine different benchmarks, in comparison to the recent SOTA compact models, in terms of FR performance and model compactness.
In a detailed comparison, different versions of our PocketNets scored SOTA performances in both, under 1M parameters and under 2M parameters, FR model categories.
Moreover, PocketNets achieved very competitive results to much larger FR models, and even outperformed them in many cases.


\section{Methodology} 
\vspace{-1mm}
This section presents the methodology leading to our proposed PocketNets solution, both the architecture design and the training paradigm. We first present the NAS process leading to the architecture of our proposed PocketNets. Then, we present our proposed multi-step knowledge distillation training paradigm.  
\vspace{-1mm}
\subsection{Towards PocketNet Architecture}
\vspace{-1mm}
Neural architecture search (NAS) automates the network design by learning the network architecture that achieves the best performance for a specific task.
NAS has proved to be a robust method in discovering and optimizing neural network architecture. Previous works \cite{cai2019proxylessnas,DBLP:conf/iclr/LiuSY19} demonstrate that the discovered network architectures by NAS do outperform handcraft-designed network architectures for different computer vision tasks. 
For our PocketNets, we opt to use differential architecture search (DARTS) algorithm \cite{DBLP:conf/iclr/LiuSY19} to search for two types of building blocks (cell) i.e. normal cell and reduce cell, which can be stacked to form the final architecture. Our choice for DARTS is based on: a) it achieved a competitive result to the SOTA NAS solutions on different image classification tasks \cite{DBLP:conf/iclr/LiuSY19}, and b) the search time for DARTS is feasible in comparison to other search methods \cite{zoph2017neural,zoph2018learning} and thus, it can be adapted to a large-scale dataset. Unlike common NAS algorithms that are applied on a small image size of a small dataset, our NAS will be learned on a large-scale face image dataset with relatively high resolution. In the following, we briefly present the DARTS algorithm.
Our goal here is not only to build an optimal architecture, but also to analyze the FR performance implications when optimizing such an architecture on a different learning task, as will be clarified later in this work.

DARTS aims at learning two types of cells: normal cell and reduce cell.
Each cell is a direct acyclic graph (DAG) that consist of N nodes. Each node ${x_i}$ is a latent representation, where $i \in [0,N]$. The operation space $O$ is a set of candidate operation e.g. convolutional layer, skip-connection, pooling layer etc. Each edge $(i,j)$ between node ${x_i}$ and ${x_j}$ is a candidate operation $o^{(i,j)} \in O$ that applies a particular transformation on ${x_i}$. Each candidate operation $o$ is weighted by the architecture parameter $a(i,j)$. An intermediate node $x_j$ is calculated as $x_j=\sum_{i<j , i \in[0,N]} o^{(i,j)} (x_i)$. 
Each cell (DAG) has two input nodes and a single output node. The two input nodes are the output of the previous two cells of the network.
The output of the last node $x_{N-1}$ i.e. the cell output, is a concatenation of all nodes in the DAG excluding the input nodes.
The candidate operation applied to $x^{(i)}$ is represented as a function $o(.)$.
The choice of a candidate operation is formulated by applying a Softmax function over the weights of all possible operations $O$:
\vspace{-3mm}
\begin{equation}
    \bar o^{(i,j)}(x) = \sum_{o\in O}\frac{\exp(\alpha_o^{(i,j)})}{\sum_{o' \in O} \exp(\alpha_{o'^{(i,j)}})}o(x),
\end{equation}
where $\alpha_o^{(i,j)}$ is a network architecture weight parameter of a candidate operation $o$. Therefore, the architecture search becomes a task of learning a set of parameters $\alpha=\{\alpha^{(i,j)}\}$.   
The learning procedure of DARTS is based on jointly learning the network architecture represented by $\alpha$ and the network weights $w$. Given  $L_{train}$ and $L_{val}$ as the train and validation loss, respectively. The learning objective of DARTS  is to find the optimal architecture represented by $\alpha^*$ that minimizes the validation loss $L_{val}(w^*,\alpha^*)$ with $w^*=\arg\min_w L_{train}(w, \alpha^*)$ as the best performing network weights on the training set.
The architecture parameters are learned using a bi-level optimization problem with $\alpha$ as the upper-level and $w$ the lower level variable:
\vspace{-1mm}
\begin{equation}
\begin{split}
    \min_\alpha L_{val}(w^*(\alpha), \alpha)\\
    s.t. \space   
    w^*(\alpha) = \arg\min_w L_{train}(w,\alpha).
\end{split}
\end{equation}
\vspace{-1mm}
The final discrete architecture is derived by setting $o^{(i,j)}=argmax_{ o \in O} \alpha^{(i,j)}_o$.
Given an input of the shape $w \times h \times c$, the output of the reduction cell is $w/2 \times h/2 \times 2c$ and the output of the normal cell is $w \times h \times c$.  The first two nodes of cell $k$ represent the output of the two previous cells $k-1$ and $k-2$.

\vspace{-1mm}
\paragraph{Search space:}
\vspace{-1mm}
PocketNet search space includes the following operations: 1)  $3 \times 3$, $5 \times5$, $7 \times 7$ depthwise separable convolutions \cite{DBLP:journals/corr/HowardZCKWWAA17} with kernel size of $\{3\times 3, 5\times 5, 7 \times7\}$, padding of $\{1, 2, 3\}$  to preserve the spatial resolution, and they have a stride of one (if applicable). 2) $1\times1$ Conv, a convolution layer with kernel size of $1\times1 $ and zero padding. 3) max pooling layer with kernel size of $3 \times3$. 4) average pooling layer with a kernel size of $3\times3$. 5) identity. 6) zero. A zero operation indicates that there is no connection between nodes.
The max and average pooling layers are followed by batch noramlization (BN) \cite{DBLP:conf/icml/IoffeS15}. 
We use Parametric Rectified Linear Unit (PReLU) \cite{DBLP:conf/iccv/HeZRS15} as the non-linearity in all convolutional operation.

\begin{table}[]
\centering
\resizebox{\linewidth}{!}{%
\begin{tabular}{|l|l|l|l|}
\hline
Operation                              & Output size         & R & Param.     \\ \hline
Conv2d(k=3,s=2,p=1),BN        & {[}64 x 56 x 56{]}  & 1 & 1856      \\ \hline
Normal-Cell 1-6                         & {[}64 x 56 x 56{]}  & 6 & 33,792    \\ \hline
Reduction-Cell 1                        & {[}128 x 28 x 28{]} & 1 & 10,688    \\ \hline
Normal-Cell 7-11                        & {[}128 x 28 x 28{]} & 5 & 92608     \\ \hline
Reduction-Cell 2                       & {[}256 x 14 x 14{]} & 1 & 35,712    \\ \hline
Normal-Cell 12-15                       & {[}256 x 14 x 14{]} & 4 & 60493,824 \\ \hline
Reduction-Cell 3                       & {[}512 x 7 x 7{]}   & 1 & 128,768   \\ \hline
PReLU, Conv2d(k=1), BN, PReLU & {[}512 x 7 x 7{]}   & 1 & 264192    \\ \hline
Conv2d(k=7,g=512), BN         & {[}512 x 1 x 1{]}   & 1 & 26112     \\ \hline
Conv2d(k=1), BN               & {[}128 x 1 x 1{]}   & 1 & 65792     \\ \hline
\end{tabular}%
}
\vspace{-3mm}
\caption{
Architecture of PocketNetS-128. Normal and reduction cells are the cells learned by DARTS on CASIA-WebFace. The table shows the number of parameters for each operation. If the operation contains a set of sub-operations (e.g. Conv2d, BN), the number of parameters is presented as the sum of parameters for all these sub-operations and multiplied by R. Column R indicates how many times the operation is repeated. The k of the convolution layer (Conv2d) refers to the kernel size, s is the stride, p is the padding, and g is the group parameter.  
}
\label{tab:pocket_arc}
\vspace{-5mm}
\end{table}

\vspace{-1mm}
\paragraph{PocketNet architecture:}
\vspace{-1mm}
We followed \cite{DBLP:conf/iclr/LiuSY19} by setting the number of nodes in all cells to $N=7$.
We apply fast down-sampling in the beginning of the network using $3 \times 3$ convolution (stride=2) followed by BN \cite{DBLP:conf/icml/IoffeS15}. 
To obtain the feature embedding of the input face image, we use global depthwise convolution \cite{DBLP:journals/corr/HowardZCKWWAA17} rather than using average pooling or fully connected layer directly before the classification layer. Our choice of using the global depthwise convolution for the embedding stage is based on: a) 
it contains fewer parameters than a fully connected layer, b) convolutional neural network (CNN) with global depth-wise convolution is more accurate than the one with average pooling for FR, as reported in previous works \cite{DBLP:conf/ccbr/ChenLGH18,DBLP:conf/icb/BoutrosDFKK21}.
The rest of the network architecture is constructed by stacking  $M$ normal cells and $3$ reduction cells at 1/3 and 2/3 of the network depth, and after the last normal cell.  We trained the NAS to optimize $\alpha_{normal}$ and $\alpha_{reduction}$ used to construct the normal and reduction cells, respectively.   

\begin{figure}[ht]
     \begin{subfigure}[b]{0.45\linewidth}
         \centering
         \includegraphics[width=\linewidth]{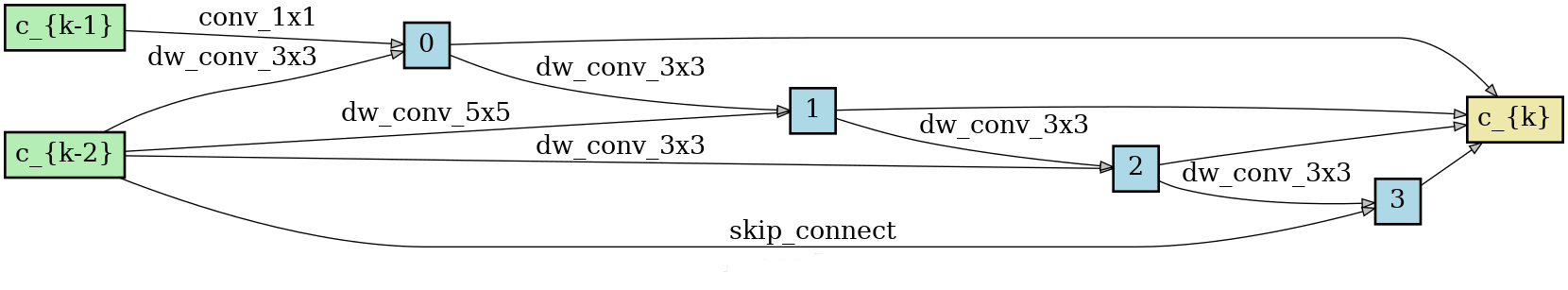}
         \caption{Normal cell learned on CASIA-WebFace.}
         \label{fig:casia_noraml}
     \end{subfigure}
      \begin{subfigure}[b]{0.45\linewidth}
         \centering
         \includegraphics[width=\linewidth]{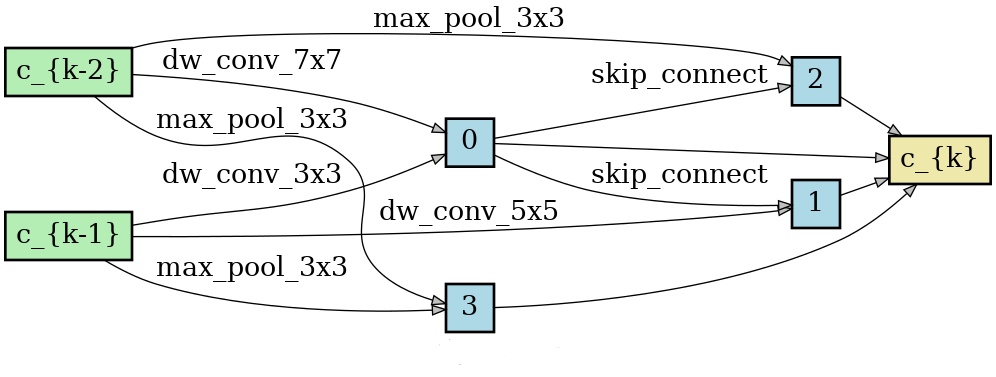}
         \caption{Reduction cell learned on CASIA-WebFace.}
         \label{fig:casia_reuction}
     \end{subfigure}
     
          \begin{subfigure}[b]{0.45\linewidth}
         \centering
         \includegraphics[width=\linewidth]{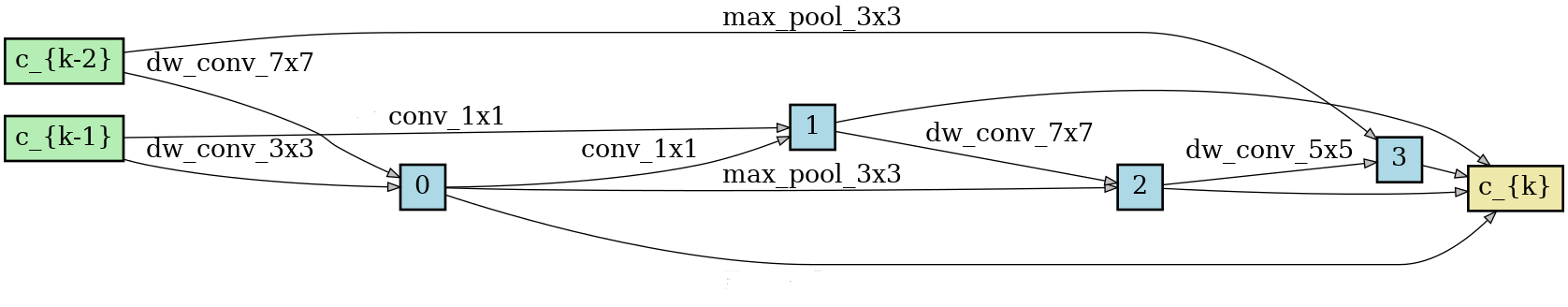}
         \caption{Normal cell learned on CIFAR-10.}
         \label{fig:cifar_noraml}
     \end{subfigure}
      \begin{subfigure}[b]{0.45\linewidth}
         \centering
         \includegraphics[width=\linewidth]{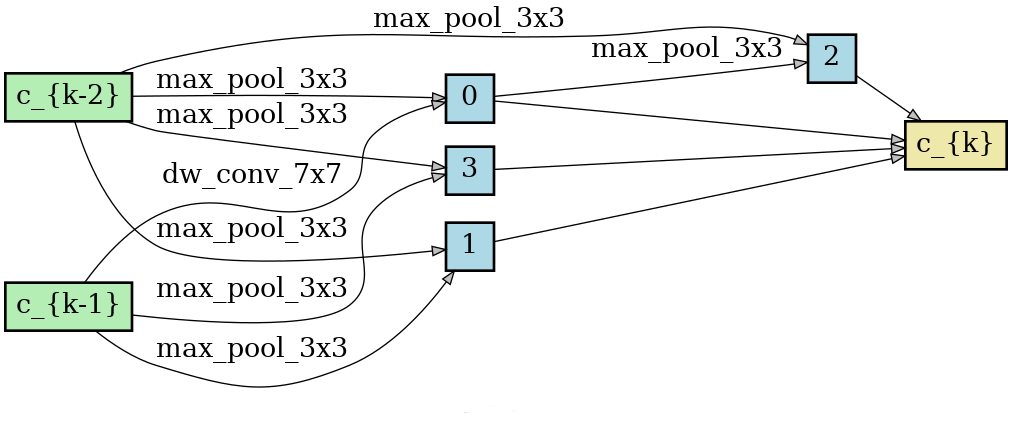}
         \caption{Reduction cell learned on CIFAR-10.}
         \label{fig:cifar_reuction}
     \end{subfigure}
       \vspace{-2mm}
        \caption{Normal and reduction cells learned by DARTS on CASIA-WebFace and CIFAR-10 datasets. 
       }
        \label{fig:cell}
\vspace{-4mm}
\end{figure}

We trained the search algorithm to learn from the CASIA-WebFace dataset \cite{DBLP:journals/corr/YiLLL14a}. Training details are presented later in Section \ref{sec:nas_exp}.
The best discovered normal and reduction cells by DARTS are shown in Figures \ref{fig:casia_noraml} and \ref{fig:casia_reuction}, respectively.
In this work, we present four architectures based on the learned cells: PocketNetS-128, PocketNetS-256, PocketNetM-128, and PocketNetM-256. The architecture of PocketNetS-128 and PocketNetS-256 (PocketNet small) are identical. Each of them contains 18 cells i.e 15 normal cells and 3 reduction cells. 
The number of feature maps (out channel) of the first layer is 64.
The only difference is the embedding size, where the embedding in PocketNetS-128 is of size 128-D and in PocketNetS-256 is of size 256-D.
Table \ref{tab:pocket_arc} presents the overall architecture of PocketNetS-128. 
PocketNetS-128 contains in total 925,632 trainable parameters and setting the embedding size to 256 increases the number of parameters in PocketNetS-256 to 991,424.
All networks use floating-point 32 and the required memory footprints are 3.7 and 3.9 MB by PocketNetS-128 and PocketNetS-256, respectively.
The main motivation for using different embedding sizes is to evaluate the effect of embedding size on the network performance and memory footprint. We also investigate a wider architecture of PocketNet by doubling the number of feature maps of the network and reducing the number of cells from 18 to 9. This result in two networks: PocketNetM-128 and PocketNetM-256 (PocketNet medium) with embedding size of 128-D and 256-D, respectively. The architecture of PocketNetM-128 is presented in Table \ref{tab:pocketM_arc}. PocketNetM-128 contains 1,686,656 parameters and PocketNetM-256 contains 1,752,448 parameters.

\begin{table}[]
\centering
\resizebox{\linewidth}{!}{%
\begin{tabular}{|l|l|l|l|}
\hline
Operation                              & Output size         & R & Param   \\ \hline
Conv2d(k=3,s=2,p=1),BN        & {[}128 x 56 x 56{]} & 1 & 3712    \\ \hline
Normal-Cell1-6                         & {[}128 x 56 x 56{]} & 3 & 56832   \\ \hline
Reduction-Cell 1                        & {[}256 x 28 x 28{]} & 1 & 35,712  \\ \hline
Normal-Cell 7-11                        & {[}256 x 28 x 28{]} & 2 & 128896  \\ \hline
Reduction-Cell 2                       & {[}512 x 14 x 14{]} & 1 & 128,768 \\ \hline
Normal-Cell 12-15                       & {[}512 x 14 x 14{]} & 1 & 227,072 \\ \hline
Reduction-Cell 3                       & {[}1024 x 7 x 7{]}  & 1 & 486,912 \\ \hline
PReLU, Conv2d(k=1), BN, PReLU & {[}512 x 7 x 7{]}   & 1 & 526848  \\ \hline
Conv2d(k=7,g=512), BN         & {[}512 x 1 x 1{]}   & 1 & 26112   \\ \hline
Conv2d(k=1), BN               & {[}128 x 1 x 1{]}   & 1 & 65792   \\ \hline
\end{tabular}%
}
\vspace{-3mm}
\caption{Architecture of PocketNetM-128. Normal and reduction cells are the cells learned by DARTS on CASIA-WebFace. The table shows the number of parameters for each operation. If the operation contains a set of sub-operations (e.g. Conv2d, BN), the number of parameters is presented as the sum of parameters for all these sub-operations and multiplied by R. Column R indicates how many times the operation is repeated.
The k of the convolution layer (Conv2d) refers to the kernel size, s is the stride, p is the padding, and g is the group parameter.  
}
\label{tab:pocketM_arc}
\vspace{-4mm}
\end{table}



\subsection{PocketNet Training Paradigm}
\label{sec:loss}
Towards the PocketNet training paradigm that incorporates our proposed multi-Step KD, we start by formulating the margin-based Softmax loss and knowledge distillation concept.
Margin-Based Softmax loss has been widely deployed in recent FR solutions \cite{deng2019arcface,DBLP:conf/cvpr/WangWZJGZL018,MagFace}. It achieved SOTA accuracy on major benchmarks \cite{deng2019arcface,martinez2021benchmarking,MagFace}. In this work, we utilize the ArcFace loss \cite{deng2019arcface} to train our PocketNets.   ArcFace loss extends over the softmax loss by manipulating the decision boundary between the classes by deploying an additive angular margin penalty on the angle between the weights of the last fully connected layer and the feature representation. Formally, ArcFace loss is defined as follow:
\vspace{-1mm}
\begin{equation}
\resizebox{\linewidth}{!}{$
    L_{Arc}=\frac{1}{M}  \sum\limits_{i \in M} - log \frac{e^{s (cos(\theta_{y_i}+m))}}{ e^{s(cos(\theta_{y_i}+m))} +\sum\limits_{j=1 , j \ne y_i}^{C}  e^{s ( cos(\theta_{j}))}},
$}\end{equation}
where $\theta_{yi}$ is the angle between the feature $f_i$ and $i-th$ class center, $y_i \in [1,C]$ (C is the number of classes), $M$ is batch size, $m$ is the margin penalty value and $s$ is scale parameter.

\vspace{-3mm}

\paragraph{Knowledge distillation (KD):} 
KD is a technique to improve the performance and generalizability of smaller models by transferring the knowledge learned by a cumbersome model (teacher) to a single small model (student) \cite{DBLP:journals/corr/HintonVD15}. 
The idea is to guide the student model to learn the relationship between different classes discovered by the teacher model that contains more complex information beyond the ground truth labels \cite{DBLP:journals/corr/HintonVD15}. The KD is originally proposed to improve the performance of a small backbone trained with SoftMax loss for a classification task \cite{DBLP:journals/corr/HintonVD15}.
However, the learning objective of the FR model is to optimize feature representations needed for face verification. 
In this work, as a step towards our proposed multi-step KD, we train our PocketNet model to learn feature representations that are similar to the ones learned by the teacher model. 
We achieve that by introducing an additional loss function (Mean squared error (MSE)) to ArcFace loss operated on the embedding layer. Formally,the $l_{mse}$ loss is defined as follows:
\vspace{-1mm}
\begin{equation}
\label{eq:mse}
    l_{mse}= \frac{1}{M}  \sum\limits_{i \in M} 1- \frac{1}{D}\Sigma_{h=1}^{D}{\Big(\Phi^S_t(x)_h -\Phi^T_t(x)_h\Big)^2}, 
\end{equation}
where $\Phi^S_t$ and $\Phi^T_t$ are the feature representations obtained from the last fully connected layer of student and teacher models, respectively, and D is the size of the feature representation. The final training loss function is defined as follow:
\vspace{-1mm}
\begin{equation}
    l_{mse}=l_{Arc} + \lambda l_{mse},
\end{equation}
where $\lambda$ is a weight parameter. The feature representations learned by the ArcFace loss are normalized. Thus, the value range of $l_{mse}$ is much small i.e. $\leq$ 0.007.
This value is very small in comparison to the ArcFace loss value (around 60 at the beginning of the training phase.) 
We set the $\lambda$ value to 100. Thus, the $l_{mse}$ contributes to the model training. 

\vspace{-3mm}
\paragraph{Multi-Step Knowledge Distillation:}
Previous works \cite{DBLP:conf/aaai/MirzadehFLLMG20,DBLP:conf/iccvw/YanZXZWS19} observed that transforming the knowledge from a very deep teacher model to a small student model is difficult when the gap in terms of network size between the teacher and the student model is large. 

In this work, we present a novel concept by relaxing this difficulty of a substantial discrepancy between teacher model and student by synchronizing the student and the teacher model during the training, without the need for transforming the knowledge to intermediate networks \cite{DBLP:conf/aaai/MirzadehFLLMG20,DBLP:conf/iccvw/YanZXZWS19}.
Our solution is designed to transfer the knowledge learned by a teacher model in a step-wise manner after each $x$ number of iterations, i.e. Multi-Step KD.  
The key idea is that the information learned by a teacher at different steps of the training phase is different from the one learned when the teacher is fully converged. 
Thus, transferring the knowledge learned by a teacher at an early stage of training is easier for a student to learn. 
Thus, at a later point when the student is converged to some degree, it can learn more complex patterns from the teacher.  
To achieve that, we first train the teacher for $I$ iterations. 
This teacher model is noted as $T1$. 
Then, we train the student model for the same number of iteration $I$ with the assistance of the teacher $T1$.
In this case, $\Phi^T$ (Equation \ref{eq:mse}) is $T1$ obtained after the first $I$ iterations.  
We choose to train the teacher for one epoch each time. 
This will give the teacher a chance to learn from the whole training dataset.  
We repeated these two steps until the teacher and student models are converged. 
To simplify the implementation, we train first the teacher model until it is converged and save the model weights after each epoch. 
Then, we train the student model with the assistance of the teacher models. 
During the student training, we load the teacher weights that correspond to the same training epoch.

\vspace{-3mm}
\begin{figure}[h!]
     \centering
     \begin{subfigure}[b]{0.85\linewidth}
         \centering
         \includegraphics[width=\linewidth]{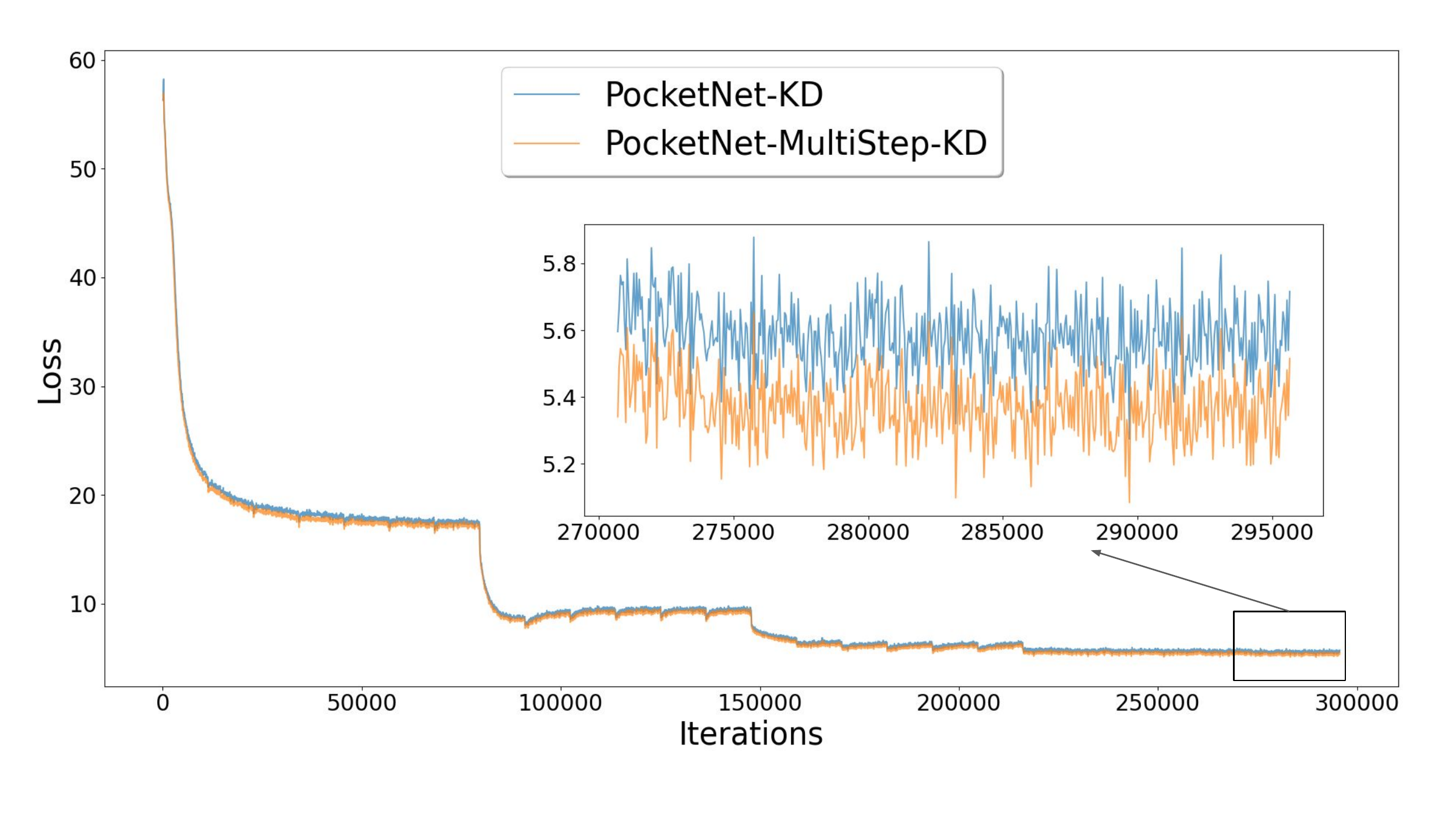}
        \caption{ArcFace loss value of the model trained with KD vs. the model trained with multi-step KD over training iterations.}
         \label{fig:loss_arc}
     \end{subfigure}
     
      \begin{subfigure}[b]{0.85\linewidth}
         \centering
         \includegraphics[width=\linewidth]{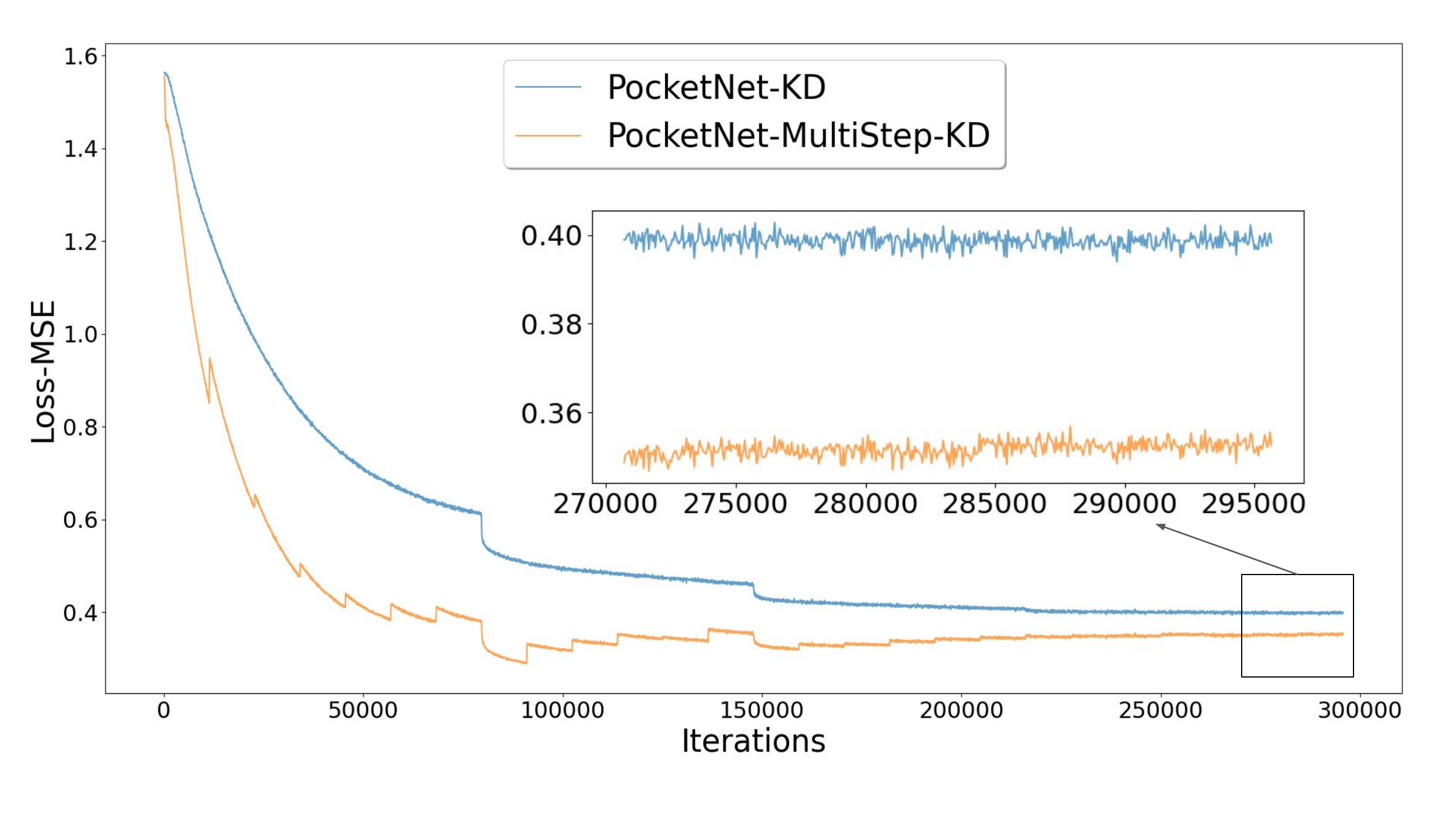}
         \caption{KD vs. multi-step KD loss values over training iterations.}
         \label{fig:loss_mse}
     \end{subfigure}
     \vspace{-2mm}
     \caption{ Effect of multi-step KD on the student model convergence. It can be noticed that multi-step KD enables the model trained with ArcFace and multi-step KD losses to better converges in comparison to the case where the model is trained with ArcFace and KD losses (Figure \ref{fig:loss_arc}). Also, it can be observed that training with multi-step KD guides the model to learn feature representations that are more similar (in comparison to KD) to the teacher ones (Figure \ref{fig:loss_mse}). These figures are based on training the PocketNetS-128 network.     }
     \label{fig:loss}
     \vspace{-5mm}
\end{figure}

\section{Experimental Setups} 
\subsection{Neural Architecture Search}
\label{sec:nas_exp}
We train the DARTS to learn the normal and reduction cells on the CASIA-Webface dataset \cite{DBLP:journals/corr/YiLLL14a}. 
CASIA-Webface consists of 494,141 face images from 10,757 different identities. We split the dataset equally into two 
parts used for training and validation. The images are pre-aligned and cropped to $120 \times 120$ for the training subset and to $112 \times 112$ for the validation subset using the Multi-task Cascaded Convolutional Networks (MTCNN) solution \cite{zhang2016joint}.
During the training phase, the training images are randomly cropped to have a fixed size of $112 \times 112$ and then randomly horizontally flipped to make the search more robust, following common practice in FR research \cite{deng2019arcface,MagFace}.  
All the training and validation images are normalized to have pixel values between -1 and 1.
We followed DARTS training setup \cite{DBLP:conf/iclr/LiuSY19} by using Stochastic Gradient Descent with the momentum of $0.9$ and weight decay of $3e-4$ to optimize the DARTS weight $w$. 
We utilize a cosine annealing strategy \cite{DBLP:conf/iclr/LoshchilovH17} to decrease the learning rate after each epoch with a minimum learning rate of $0.004$.
We set the batch size to $128$ and the initial learning rate to $0.1$. 
For $\alpha$ optimization, we use similar setup to DARTS \cite{DBLP:conf/iclr/LiuSY19} by using Adam optimizer with momentum $\beta=(0.5, 0.999)$ and weight decay of $1e-3$. We set the initial learning rate for Adam optimizer to $0.0012$. 
The initial channel size is set to $64$ and the number of nodes in each cell is set to $8$. 
We use a batch size of $128$ and train DARTS for 50 epochs. 
These configurations are chosen to enable DARTS training on available GPUs.
All training codes are implemented in Pytorch \cite{NEURIPS2019_9015} and trained on 6 NVIDIA GeForce RTX 2080 Ti (11GB) GPUs. The training lasted 2274 hours.
We additionally conducted an additional experiment on CIFAR-10 \cite{Krizhevsky09learningmultiple} as a NAS domain ablation study for this work.
The CIFAR-10 is a commonly used dataset for object detection and image classification tasks consisting of 60000 images (of the size $32 \times32$) of 10 classes.
We split CIFAR-10 equally into two parts: training and validation subsets. We run the DARTS search using the exact configurations described previously in this section to learn on the CIFAR-10 dataset. 
The training lasted around 30 hours on 6 NVIDIA GeForce RTX 2080 Ti (11GB) GPUs.

\begin{figure*}[ht!]
     \centering
     \begin{subfigure}[b]{0.195\linewidth}
         \centering
         \includegraphics[width=\linewidth]{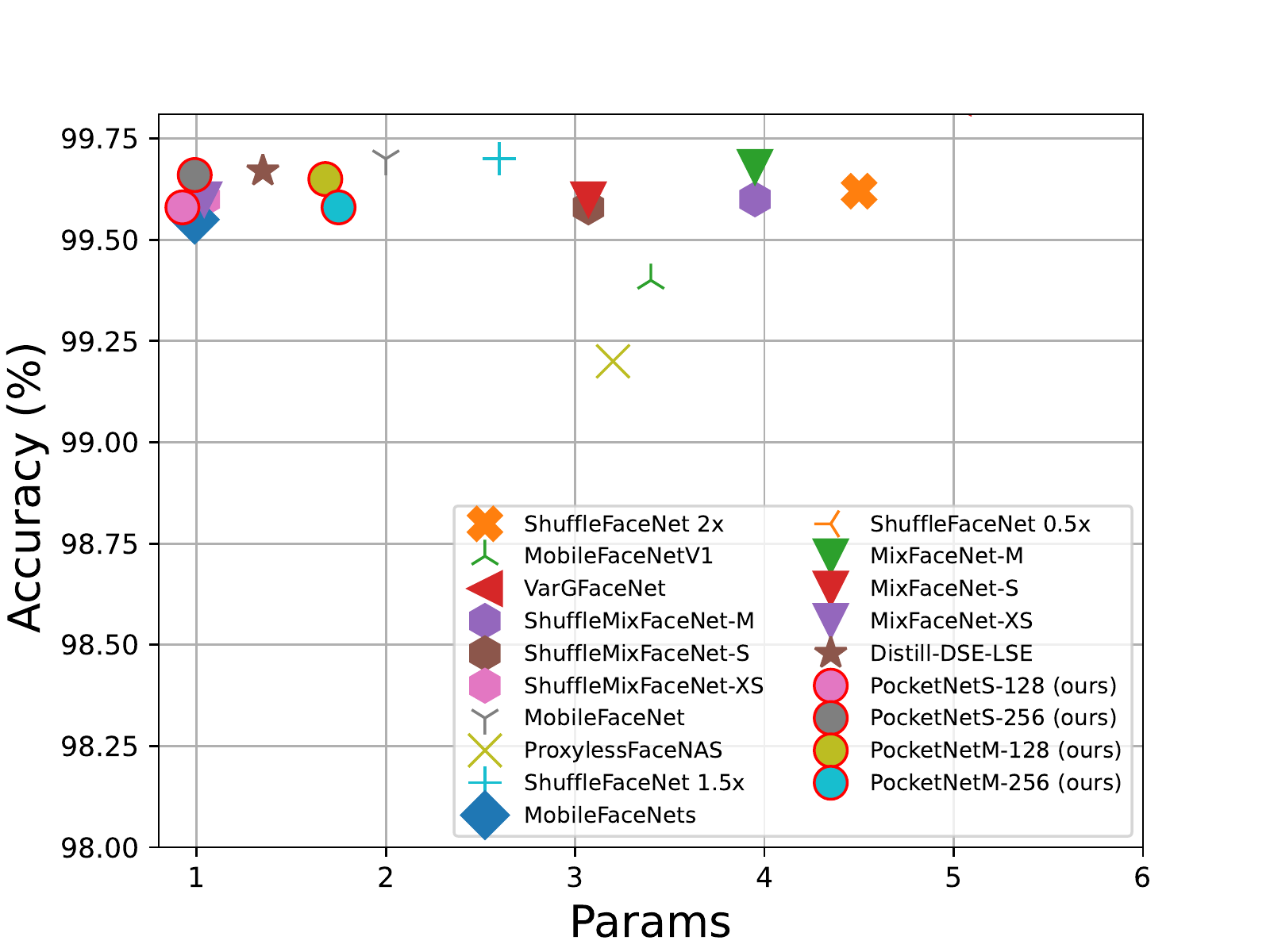}
         \caption{LFW}
         \label{fig:lfw}
     \end{subfigure}
     \begin{subfigure}[b]{0.195\linewidth}
         \centering
         \includegraphics[width=\linewidth]{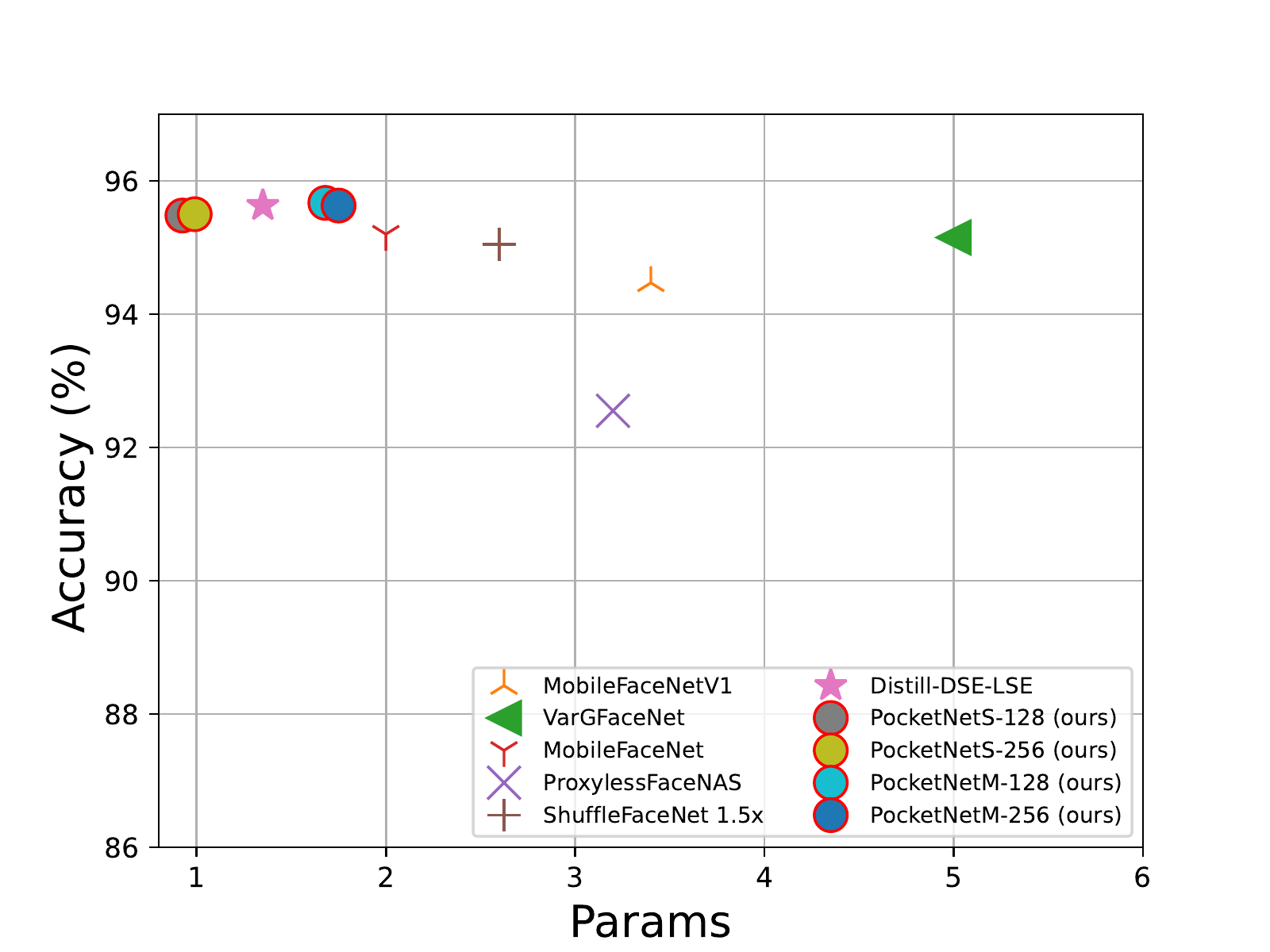}
         \caption{CA-LFW}
         \label{fig:calfw}
     \end{subfigure}
     \begin{subfigure}[b]{0.195\linewidth}
         \centering
         \includegraphics[width=\linewidth]{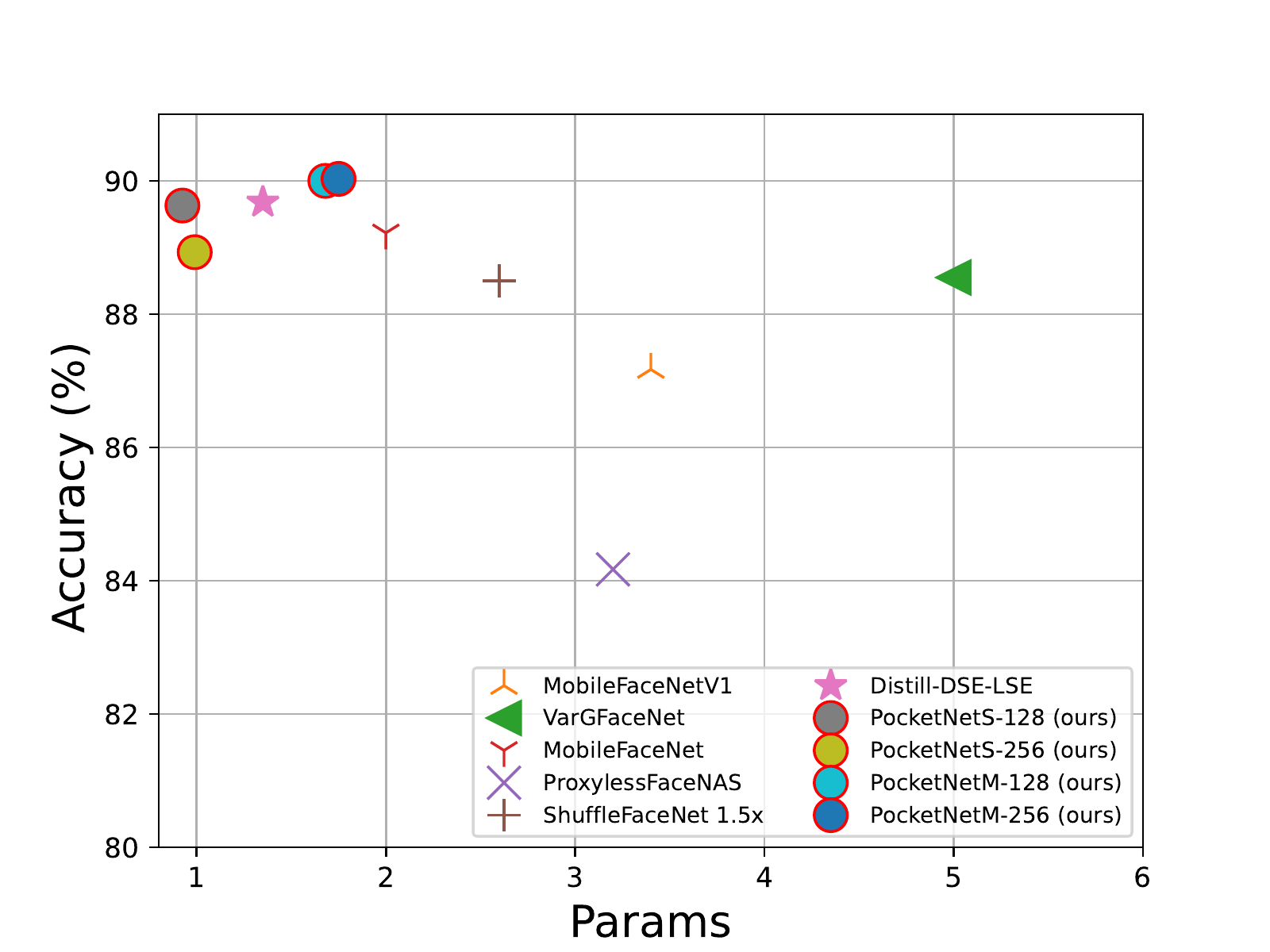}
         \caption{CP-LFW}
         \label{fig:cplfw}
     \end{subfigure}
     \begin{subfigure}[b]{0.195\linewidth}
         \centering
         \includegraphics[width=\linewidth]{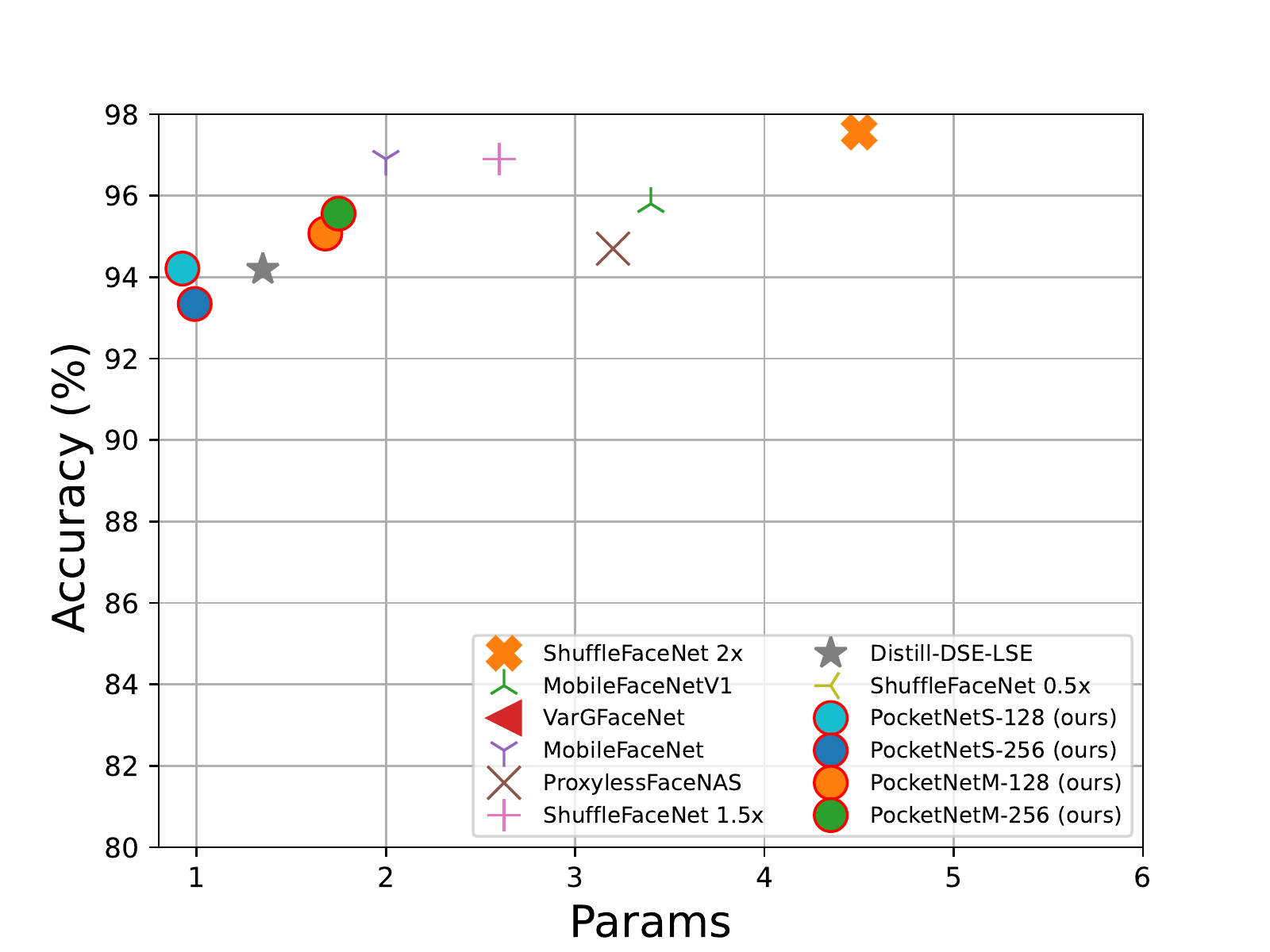}
         \caption{CFP-FP}
         \label{fig:cfp}
     \end{subfigure}
      \begin{subfigure}[b]{0.195\linewidth}
         \centering
         \includegraphics[width=\linewidth]{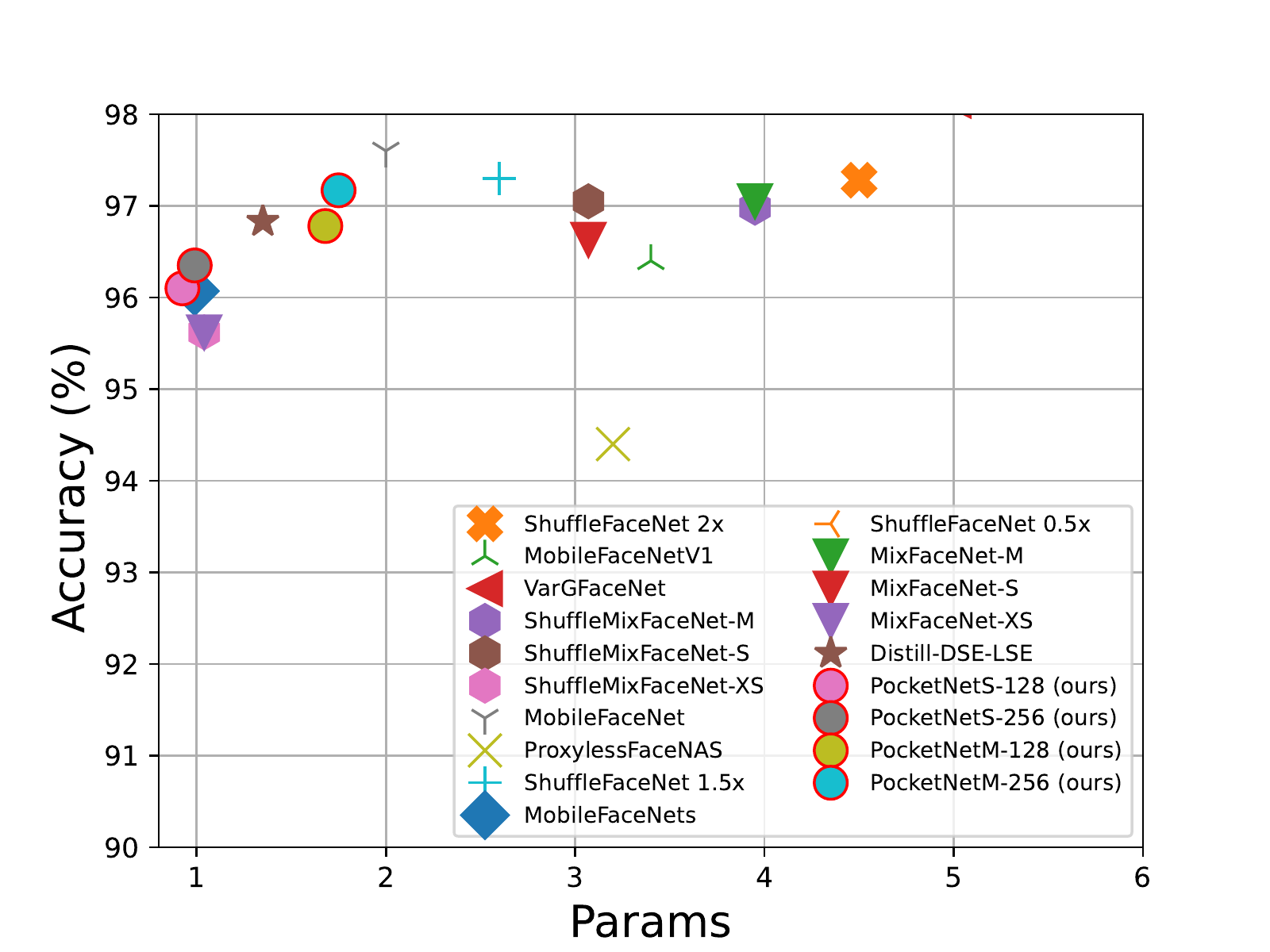}
         \caption{AgeDB-30}
         \label{fig:agedb}
     \end{subfigure}
     
     \begin{subfigure}[b]{0.195\linewidth}
         \centering
         \includegraphics[width=\linewidth]{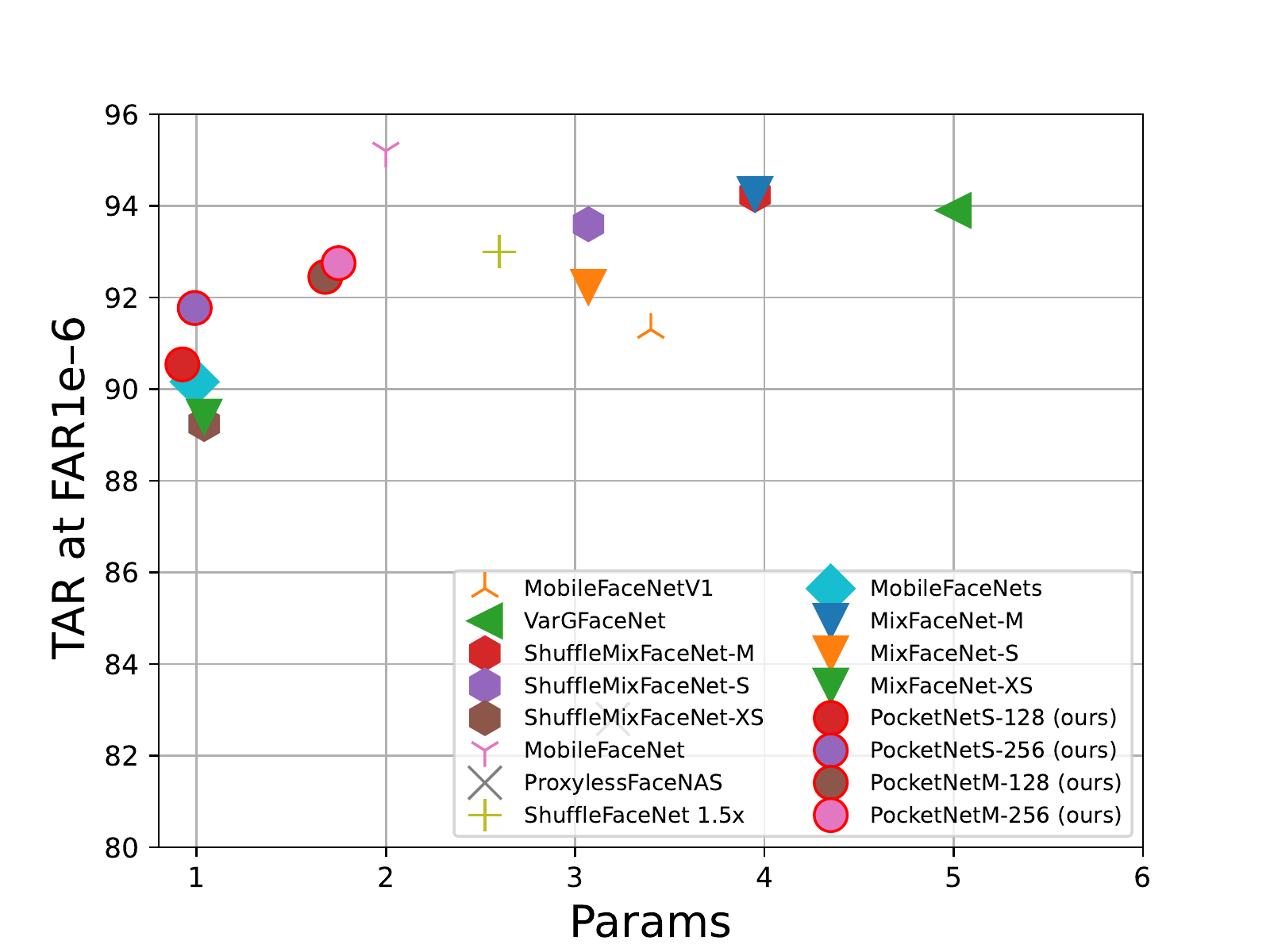}
         \caption{MegaFace}
         \label{fig:megaface}
     \end{subfigure}
    \begin{subfigure}[b]{0.195\linewidth}
         \centering
         \includegraphics[width=\linewidth]{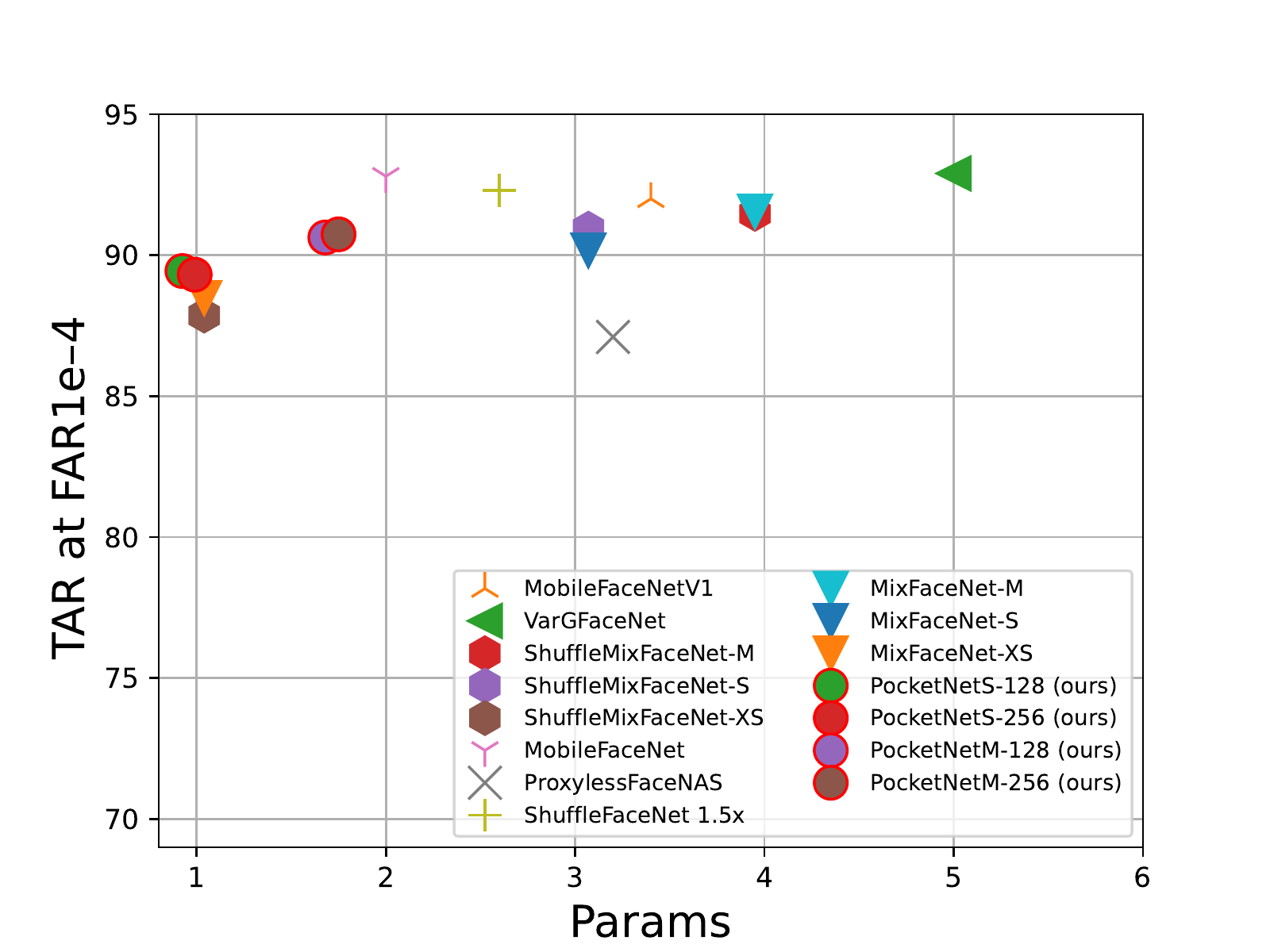}
         \caption{IJB-B}
         \label{fig:ijbb}
     \end{subfigure}
      \begin{subfigure}[b]{0.195\linewidth}
         \centering
         \includegraphics[width=\linewidth]{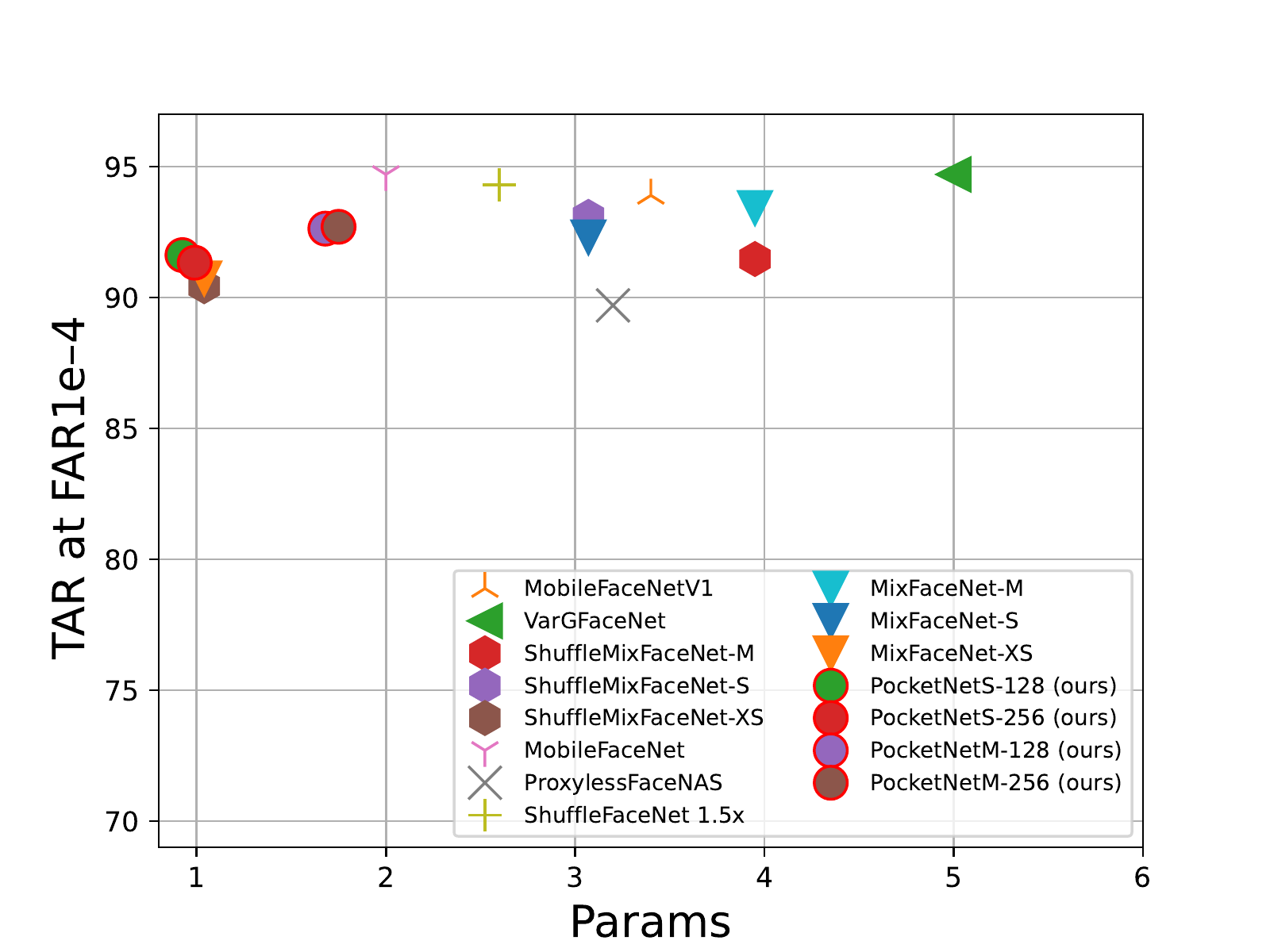}
         \caption{IJB-C}
         \label{fig:ijbc}
     \end{subfigure}
     \begin{subfigure}[b]{0.195\linewidth}
         \centering
         \includegraphics[width=\linewidth]{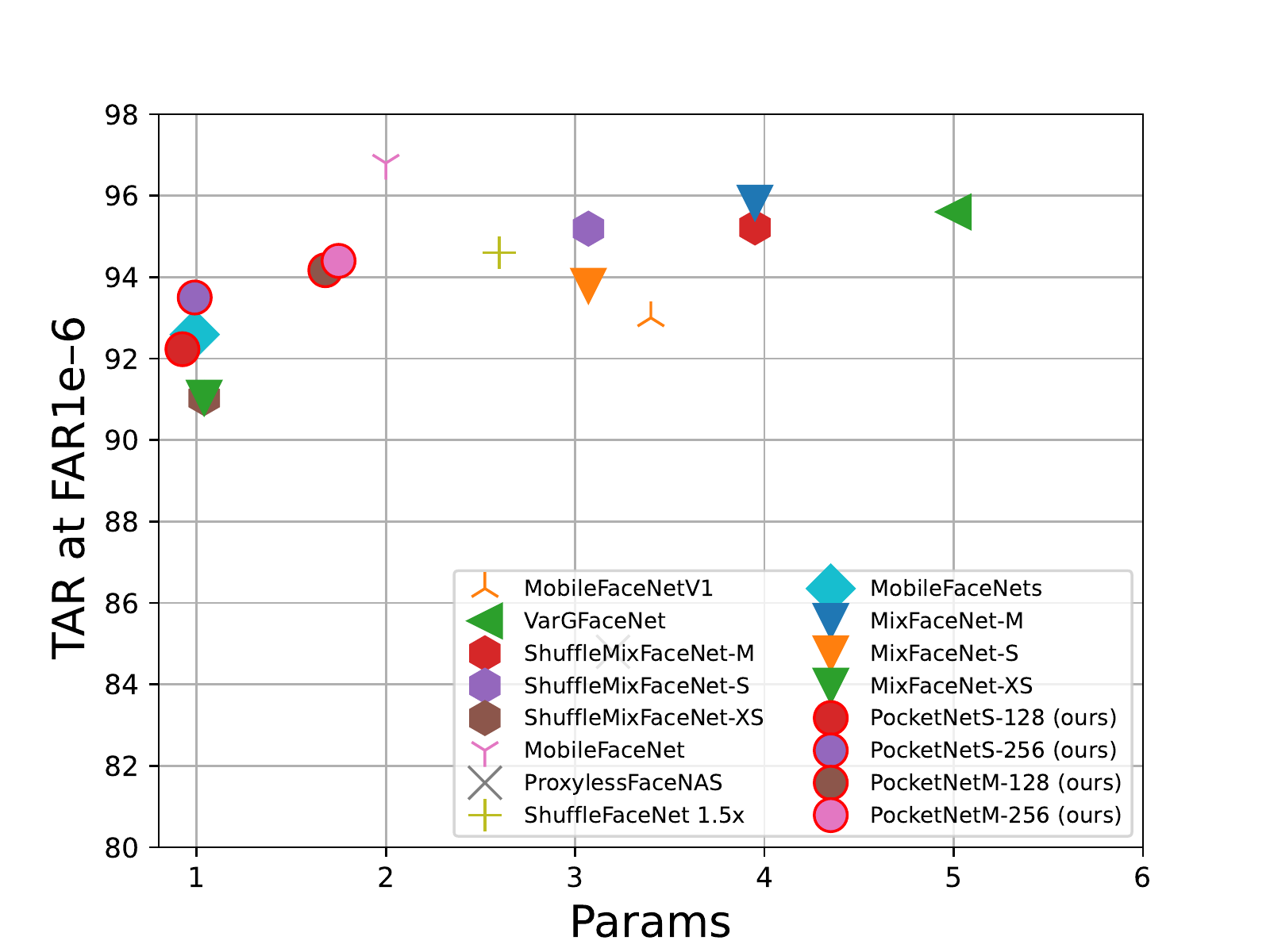}
         \caption{MegaFace (R)}
         \label{fig:megafacer}
     \end{subfigure}
        \caption{  Number of parameters (in millions) vs. performance on LFW (accuracy), CA-LFW (accuracy), CP-LFW (accuracy), CFP-FP (accuracy), AgeDB-30 (accuracy),  MegaFace (TAR at FAR1e-6), IJB-B (TAR at FAR1e-4),  IJB-C (TAR at FAR1e-4) and  MegaFace (R), (TAR at FAR1e-6).
        Our PocketNets are marked with circle marker and red edge color and are placed repeatedly in the top left corner, proving a SOTA trade-off between FR performance and compactness.
       }
        \label{fig:params}
\vspace{-3mm}
\end{figure*}


\subsection{Face Recognition models and training}
\label{sec:train_setup}
Based on the normal and reduction cells learned by DARTS on CASIA-WebFace \cite{DBLP:journals/corr/YiLLL14a}, we trained three instances of PocketNetS-128. 
The first instance (noted as PocketNetS-128 (no KD))  is only trained with ArcFace loss described in Section \ref{sec:loss}. 
The second instance (noted as PocketNetS-128 (KD)) is trained with ArcFace loss with KD. 
The third instance is trained with ArcFace loss along with our proposed multi-step KD (noted as PocketNetS-128 (multi-step KD)).
These three instances are used in our ablation study towards the proposed multi-step KD.
On the other hand, based on the normal and reduction cells learned on CIFAR-10 \cite{Krizhevsky09learningmultiple} (object classification domain), we train another model based on these cells, noted as DartFaceNet-128 (no KD).
This training is used as an ablation study to analyze the effect of training dataset sources on the neural architecture search algorithm by comparing its FR performance to its direct counterpart PocketNetS-128 (no KD).

Additionally, as detailed earlier, we trained four instances of PocketNets: PocketNetS-128, PocketNetS-256, PocketNetM-128, and PocketNetM-256 to compare our proposed PocketNets with the recent compact FR models proposed in the literature on different levels of compactness. 
All these models are trained with ArcFace loss along with our proposed multi-step KD. 
To enable KD  multi-step solutions, we trained two instances of the ResNet-100 model with embedding sizes of $128-D$ and $256-D$. The ResNet-100(128) is used as a teacher for PocketNetS-128 and PocketNetM-128, while  ResNet-100(256) is used as a teacher for PocketNetS-256 and PocketNetM-256.

We use the MS1MV2 dataset \cite{deng2019arcface} to train all the investigated FR models in this work.   
The MS1MV2 is a refined version \cite{deng2019arcface} of the MS-Celeb-1M \cite{DBLP:conf/eccv/GuoZHHG16} containing 5.8M images of 85K identities.
We follow the common setting \cite{deng2019arcface} to set the scale parameter $s$ to 64 and margin value of ArcFace loss to 0.5.  
We set the mini-batch size to 512 and train our models on a single Linux machine (Ubuntu 20.04.2 LTS) with Intel(R) Xeon(R) Gold 5218 CPU  2.30GHz, 512 G RAM, and 4 Nvidia GeForce RTX 6000 GPUs. 
The proposed models in this paper are implemented using Pytorch \cite{NEURIPS2019_9015}.
All FR models are trained with Stochastic Gradient Descent (SGD) optimizer with an initial learning rate of 1e-1. 
We set the momentum to 0.9 and the weight decay to 5e-4. 
The learning rate is divided by 10 at 80k, 140k, 210k, and 280k training iterations. 
The total number of training iteration is 295K.
During the training, we use random horizontal flipping with a probability of 0.5 for data augmentation.
The networks are trained (and evaluated) on images of the size $112 \times 112 \times 3$, with pixel values between -1 and 1.
These images are aligned and cropped using the Multi-task Cascaded Convolutional Networks (MTCNN)  \cite{zhang2016joint}, following \cite{deng2019arcface}. 


\subsection{Evaluation benchmarks and metrics}
\label{sec:metric}
We evaluate our PocketNets and build a comparison to SOTA based on 9 benchmarks detailed in this section.
The considered evaluation benchmarks are Labeled Faces in the Wild (LFW)  \cite{LFWTech}, Cross-age LFW (CA-LFW)  \cite{DBLP:journals/corr/abs-1708-08197},  Cross-Pose LFW (CP-LFW) \cite{CPLFWTech}, Celebrities in Frontal-Profile in the Wild (CFP-FP) \cite{DBLP:conf/wacv/SenguptaCCPCJ16}, AgeDB-30 \cite{DBLP:conf/cvpr/MoschoglouPSDKZ17}, IARPA Janus Benchmark-B (IJB-B) \cite{DBLP:conf/cvpr/WhitelamTBMAMKJ17}, IARPA Janus Benchmark–C (IJB-C) \cite{DBLP:conf/icb/MazeADKMO0NACG18}, MegaFace \cite{DBLP:conf/cvpr/Kemelmacher-Shlizerman16}, and MegaFace (R) \cite{deng2019arcface,DBLP:conf/cvpr/Kemelmacher-Shlizerman16}.
We acknowledge the evaluation metrics in the ISO/IEC 19795-1 \cite{mansfield2006information} standard, however, for comparability, we follow the evaluation metrics defined in the utilized benchmarks as follows: LFW (accuracy), CA-LFW (accuracy), CP-LFW (accuracy), CFP-FP (accuracy), AgeDB-30 (accuracy),  MegaFace (Rank-1 identification rate and true acceptance rates (TAR) at false acceptance rates (FAR) of 1e-6), IJB-B (TAR at FAR1e-4),  IJB-C (TAR at FAR1e-4) and  MegaFace (R), (Rank-1 identification rate and TAR at FAR1e-6). 


\vspace{-3mm}
\section{Ablation Study} 
\vspace{-2mm}
This section presents two ablation studies addressing the two main aspects of our design of the PocketNets solution. 

\vspace{-3mm}
\paragraph{Ablation Study on NAS training dataset source:}
We trained two different instances of DARTS search algorithm to learn from CASIA-WebFace \cite{DBLP:journals/corr/YiLLL14a} (face images) and CIFAR-10 \cite{Krizhevsky09learningmultiple} (animals, cars, etc.), respectively.  
Figure \ref{fig:cell} presents the normal and reduction cells learned on CASIA-WebFace and CIFAR-10,  used to build our PocketNetS-128 (no KD) and the  DartFaceNetS-129 (no KD), respectively.
These networks share the same structure including the embedding stage and the number of cells. These networks are trained using the exact training setup described in Section \ref{sec:train_setup}.
DartFaceNetS-128 (no KD) contains 885,184 parameters with 620.9286 MFLOPs. PocketNetS-128 (no KD) contains 925,632 parameters with 587.11 MFLOPs. Table \ref{tab:kd}  presents the achieved performance by PocketNetS-128 (no KD) and DartFaceNetS-128 (no KD) on nine different benchmarks. It can be clearly noticed that PocketNetS-128 (no KD) outperformed DartFaceNetS-128 (no KD) with an obvious margin on all considered benchmarks.  The demonstrates that utilizing neural network architecture designed for common computer vision tasks leads to sub-optimal performance when it is used for the FR. It also supports our choice for training NAS to learn from a face image dataset and points out that FR does require face-specific architecture design.

\vspace{-3mm}
\paragraph{Ablation Study on Multi-Step KD:}
Here, we prove the benefit of introducing our Multi-step KD training process on the PocketNet FR performance. This step-wise ablation study first looks into the advances provided by the KD training in comparison to training with no KD, proving the advancement achieved by our multi-step KD in comparison to KD.
Introducing KD to the PocketNet training phase improved the verification performances on all evaluation benchmarks by comparing PocketNetS-128 (no KD) to PocketNetS-128 (KD), ass observed in Table \ref{tab:kd}.
PocketNetS-128 (no KD) is trained only with ArcFace loss, while  PocketNetS-128 (KD) is trained with ArcFace along with KD from the ResNet-100 model. 
When  PocketNetS-128 is trained with ArcFace along with our multi-step KD (i.e. PocketNetS-128 (multi-step KD)), the achieved verification performance improved in eight out of nine different benchmarks in comparison to PocketNetS-128 (KD) (Table \ref{tab:kd}), empirically proving the benefit of our multi-step KD.
We also investigated the competence of our proposed multi-step KD on improving the model convergence.
Figure \ref{fig:loss_arc} presents a comparison between ArcFace loss values of PocketNetS-128 (KD) and PocketNetS-128 (multi-step KD). 
It can be noticed that multi-step KD improved the model convergence. 
Also, our multi-step KD enhanced the similarity between the feature representation of the teacher model and the student model. 
This observation is seen in Figure \ref{fig:loss_mse} where the MSE values of PocketNetS-128 (multi-step KD) is smaller than the one of PocketNetS-128 (KD).

\begin{table*}[ht!]
\centering
\resizebox{\textwidth}{!}{%
\begin{tabular}{|c|l|l|c|c|c|c|c|c|c|c|c|c|c|}
\hline
 &
   &
   &
   &
   &
   &
  \cellcolor[HTML]{FFFFFF}{\color[HTML]{000000} } &
   &
   &
   &
  \multicolumn{2}{c|}{MegaFace} &
  \multicolumn{2}{c|}{MegaFace (R)} \\ \cline{11-14} 
\multirow{-2}{*}{Model} &
  \multirow{-2}{*}{Param. (M)} &
  \multirow{-2}{*}{MFLOPs} &
  \multirow{-2}{*}{\begin{tabular}[c]{@{}c@{}}LFW\\ (\%)\end{tabular}} &
  \multirow{-2}{*}{\begin{tabular}[c]{@{}c@{}}CA-LFW\\ (\%)\end{tabular}} &
  \multirow{-2}{*}{\begin{tabular}[c]{@{}c@{}}CP-LFW\\ (\%)\end{tabular}} &
  \multirow{-2}{*}{\cellcolor[HTML]{FFFFFF}{\color[HTML]{000000} \begin{tabular}[c]{@{}c@{}}CFP-FP\\ (\%)\end{tabular}}} &
  \multirow{-2}{*}{\begin{tabular}[c]{@{}c@{}}AgeDB-30\\ (\%)\end{tabular}} &
  \multirow{-2}{*}{\begin{tabular}[c]{@{}c@{}}IJB-B\\ (\%)\end{tabular}} &
  \multirow{-2}{*}{\begin{tabular}[c]{@{}c@{}}IJB-C\\ (\%)\end{tabular}} &
  Rank-1(\%) &
  Ver.(\%) &
  Rank-1(\%) &
  Ver.(\%) \\ \hline
ResNet100-128  - Teacher &
  55.52 &
  24192.51 &
  99.83 &
  96.16 &
  93.1 &
  98.64 &
  98.3 &
  94.72 &
  96.08 &
  80.55 &
  97.13 &
  98.36 &
  98.66 \\ \hline \hline
DartFaceNetS-128 (no KD)&
  0.89 &
  620.9 &
  99.26 &
  94.98 &
  88.5 &
  93.18 &
  95.23 &
  87.89 &
  90.5 &
  73.44 &
  87.65 &
  87.99 &
  89.42 \\ \hline
PocketNetS-128 (no KD) &
  0.925 &
  587.11 &
  99.5 &
  95.01 &
  88.93 &
  93.78 &
  95.88 &
  88.29 &
  90.79 &
  74.42 &
  88.99 &
  89.46 &
  90.67 \\ \hline
PocketNetS-128 - KD &
  0.925 &
  587.11 &
  99.55 &
  95.15 &
  89.13 &
  93.82 &
  \textbf{96.50} &
  89.23 &
  91.47 &
  75.22 &
  90.21 &
  90.72 &
  92.04 \\ \hline
PocketNetS-128 - multi-step KD &
  0.925 &
  587.11 &
  \textbf{99.58} &
  \textbf{95.48} &
  \textbf{89.63} &
  \textbf{94.21} &
  96.10 &
  \textbf{89.44} &
  \textbf{91.62} &
  \textbf{75.81} &
  \textbf{90.54} &
  \textbf{91.22 }&
  \textbf{92.23} \\ \hline
\end{tabular}%
}
\vspace{-2mm}
\caption{Comparative evaluation results of ResNet100-128, DartFaceNetS-128 (no KD), PocketNetS-128 (no KD), PocketNetS-128 KD, and PocketNetS-128 multi-step KD on different evaluation benchmarks. The results are reported based on the evaluation metric described in Section \ref{sec:metric}.  ResNet100-128, DartFaceNetS-128 (no KD) and PocketNetS-128 (no KD) are trained with ArcFace loss.   PocketNetS-128 KD is trained with ArcFace loss with KD from teacher model (ResNet100-128).  PocketNetS-128 multi-step KD is trained with ArcFace loss with multi-step KD from teacher model (ResNet100-128). PocketNetS-128 (no KD) performed better than the DartFaceNetS-128 (no KD), proving the sanity of designing FR-specific architecture. PocketNetS-128 multi-step KD performes better than PocketNetS-128 (no KD) and PocketNetS-128 KD, proving the benefits of the proposed multi-step KD. }
\label{tab:kd}
\vspace{-2mm}
\end{table*}

\begin{table*}[h]
\centering
\resizebox{\linewidth}{!}{%
\begin{tabular}{|c|c|c|c|c|c|c|c|c|c|c|c|c|c|} 
\hline
\multirow{2}{*}{Model} & \multirow{2}{*}{Params.(M)} & \multirow{2}{*}{MFLOPs} & \multirow{2}{*}{\begin{tabular}[c]{@{}c@{}}LFW \\ (\%)\end{tabular}} 
& \multirow{2}{*}{\begin{tabular}[c]{@{}c@{}}CA-LFW\\ (\%)\end{tabular}} & \multirow{2}{*}{\begin{tabular}[c]{@{}c@{}}CP-LFW\\ (\%)\end{tabular}} & \multirow{2}{*}{\begin{tabular}[c]{@{}c@{}}CFP-FP\\ (\%)\end{tabular}} & \multirow{2}{*}{\begin{tabular}[c]{@{}c@{}}AgeDB-30\\ (\%)\end{tabular}} & \multirow{2}{*}{\begin{tabular}[c]{@{}c@{}}IJB-B\\ (\%)\end{tabular}} & \multirow{2}{*}{\begin{tabular}[c]{@{}c@{}}IJB-C\\ (\%)\end{tabular}} & \multicolumn{2}{c|}{MegaFace} & \multicolumn{2}{c|}{MegaFace(R)}  \\ 
\cline{11-14}
                       &                             &                         &                                                                      &                         &                                                                        &                                                                        &                                                                          &                                                                       &                                                                       & Rank-1 (\%) & Ver. (\%)                 & Rank-1 (\%) & Ver. (\%)                     \\ 
\hline
VarGFaceNet  \cite{DBLP:conf/iccvw/YanZXZWS19,martinez2021benchmarking}          & 5.0                         & 1022                    & 99.85                                                                & 95.15                   & 88.55                                                                  & 98.50                                                                  & 98.15                                                                    & 92.9                                                                  & 94.7                                                                  & 78.2   & 93.9                 & 94.9   & 95.6                     \\ 
\hline
ShuffleFaceNet 2×  \cite{DBLP:conf/iccvw/Martinez-DiazLV19}     & 4.5                         & 1050                    & 99.62                                                                & -                       & -                                                                      & 97.56                                                                  & 97.28                                                                    & -                                                                     & -                                                                     & -      & -                    & -      & -                        \\ 
\hline
MixFaceNet-M \cite{DBLP:conf/icb/BoutrosDFKK21}          & 3.95                        & 626.1                   & 99.68                                                                & -                       & -                                                                      & -                                                                      & 97.05                                                                    & 91.55                                                                 & 93.42                                                                 & 78.20  & 94.26                & 94.95  & 95.83                    \\ 
\hline
ShuffleMixFaceNet-M  \cite{DBLP:conf/icb/BoutrosDFKK21}   & 3.95                        & 626.1                   & 99.60                                                                & -                       & -                                                                      & -                                                                      & 96.98                                                                    & 91.47                                                                 & 91.47                                                                 & 78.13  & 94.24                & 94.64  & 95.22                    \\ 
\hline
MobileFaceNetV1  \cite{martinez2021benchmarking}      & 3.4                         & 1100                    & 99.4                                                                 & 94.47                   & 87.17                                                                  & 95.8                                                                   & 96.4                                                                     & 92.0                                                                  & 93.9                                                                  & 76.0   & 91.3                 & 91.7   & 93.0                     \\ 
\hline
ProxylessFaceNAS  \cite{martinez2021benchmarking}     & 3.2                         & 900                     & 99.2                                                                 & 92.55                   & 84.17                                                                  & 94.7                                                                   & 94.4                                                                     & 87.1                                                                  & 89.7                                                                  & 69.7   & 82.8                 & 82.1   & 84.8                     \\ 
\hline
MixFaceNet-S    \cite{DBLP:conf/icb/BoutrosDFKK21}        & 3.07                        & 451.7                   & 99.6                                                                 & -                       & -                                                                      & -                                                                      & 96.63                                                                    & 90.17                                                                 & 92.30                                                                 & 76.49  & 92.23                & 92.67  & 93.79                    \\ 
\hline
ShuffleMixFaceNet-S \cite{DBLP:conf/icb/BoutrosDFKK21}    & 3.07                        & 451.7                   & 99.58                                                                & -                       & -                                                                      & -                                                                      & 97.05                                                                    & 90.94                                                                 & 93.08                                                                 & 77.41  & 93.60                & 94.07  & 95.19                    \\ 
\hline
ShuffleFaceNet 1.5x  \cite{DBLP:conf/iccvw/Martinez-DiazLV19,martinez2021benchmarking}   & 2.6                         & 577.5                   & 99.7                                                                 & 95.05                   & 88.50                                                                  & 96.9                                                                   & 97.3                                                                     & 92.3                                                                  & 94.3                                                                  & 77.4   & 93.0                 & 94.1   & 94.6                     \\ 
\hline
MobileFaceNet  \cite{martinez2021benchmarking}        & 2.0                         & 933                     & 99.7                                                                 & 95.2                    & 89.22                                                                  & 96.9                                                                   & 97.6                                                                     & 92.8                                                                  & 94.7                                                                  & 79.3   & 95.2                 & 95.8   & 96.8                     \\ 
\hline \hline
PocketNetM-256 (Ours)  & 1.75                        & 1099.15                 & 99.58                                                                & 95.63                   & 90.03                                                                  & 95.66                                                                  & 97.17                                                                    & 90.74                                                                 & 92.70                                                                 & 78.23  & 92.75                & 94.13  & 94.40                    \\ 
\hline
PocketNetM-128 (Ours)  & 1.68                        & 1099.02                 & 99.65                                                                & 95.67                   & 90.00                                                                  & 95.07                                                                  & 96.78                                                                    & 90.63                                                                 & 92.63                                                                 & 76.49  & 92.45                & 92.77  & 94.17                    \\ 
\hline
Distill-DSE-LSE  \cite{DBLP:journals/information/LiuZC21}          & 1.35  & -       & 99.67 & 95.63 & 89.68 & 94.19 & 96.83 &   -    &      - &   -    &   -    &    -   &       \\ \hline

MixFaceNet-XS   \cite{DBLP:conf/icb/BoutrosDFKK21}       & 1.04                        & 161.9                   & 99.60                                                                & -                       & -                                                                      & -                                                                      & 95.85                                                                    & 88.48                                                                 & 90.73                                                                 & 74.18  & 89.40                & 89.35  & 91.04                    \\ 
\hline
ShuffleMixFaceNet-XS  \cite{DBLP:conf/icb/BoutrosDFKK21}  & 1.04                        & 161.9                   & 99.53                                                                & -                       & -                                                                      & -                                                                      & 95.62                                                                    & 87.86                                                                 & 90.43                                                                 & 73.85  & 89.24                & 88.823 & 91.03                    \\ 
\hline \hline
MobileFaceNets   \cite{DBLP:conf/ccbr/ChenLGH18}       & 0.99                        & 439.8                   & 99.55                                                                & -                       & -                                                                      & -                                                                      & 96.07                                                                    & -                                                                     & -                                                                     & -      & 90.16                & -      & 92.59                    \\ 
\hline
PocketNetS-256 (Ours)  & 0.99                        & 587.24                  & 99.66                                                                & 95.50                   & 88.93                                                                  & 93.34                                                                  & 96.35                                                                    & 89.31                                                                 & 91.33                                                                 & 76.53  & 91.77                & 92.29  & 93.5                     \\ 
\hline
PocketNetS-128 (Ours)  & 0.92                      & 587.11                  & 99.58                                                                & 95.48                   & 89.63                                                                  & 94.21                                                                  & 96.10                                                                    & 89.44                                                                 & 91.62                                                                 & 75.81  & 90.54                & 91.22  & 92.23                    \\ 
\hline
ShuffleFaceNet 0.5x  \cite{DBLP:conf/iccvw/Martinez-DiazLV19}   & 0.5                         & 66.9                    & 99.23                                                                & -                       & -                                                                      & 92.59                                                                  & 93.22                                                                    & -                                                                     & -                                                                     & -      & -                    & -      & -                        \\
\hline
\end{tabular}}
\vspace{-2mm}
\caption{ The achieved results on 9 benchmarks. The results are reported in \% based on the evaluation metric described in Section \ref{sec:metric}.
The models are ordered based on the number of parameters.
Our PoacketNetS-128 and PocketNetS-256 consistently extend the SOTA performance on all evaluation benchmarks for the models that have less than 1M parameters. Our PoacketNetM-128 and PocketNetM-256 also achieved SOTA performances for models that have less than 2M parameters. Additionally, they achieved very competitive results to larger models that have between 2 and 5M parameters.
All decimal points are provided as reported in the respective works.}
\label{tab:verification}
\vspace{-4mm}
\end{table*}

\vspace{-2mm}
\section{Experimental results} 
\label{sec:res}
Table \ref{tab:verification} presents the achieved FR results by our PocketNets on all evaluation benchmarks. It also presents a comparison between our proposed PocksetNets and the recent compact models proposed in the literature. 
The presented models are ordered in groups based on the number of parameters (compactness). The first part of Table  \ref{tab:verification} presents the achieved result by the models that have between 2 and 5M trainable parameters, while the second and third parts present the results for the models with less than 2M and less than 1M trainable parameters, respectively.

Our PocketNetS-128 (0.92M parameters) and PocketNetS-256 (0.99M parameters) outperformed all models that have less than 1M parameters. 
With 10\% less parameter than MobileFaceNets \cite{DBLP:conf/ccbr/ChenLGH18}, PocketNetS-128 outperformed MobileFaceNets on all considered benchmarks.
Also, PocketNetS-128 and PocketNetS-256 achieved competitive results to other deeper models that contain 4 or 5 times more parameters than PocketNets.
For example, PocketNetS-128 outperformed VarGFaceNet (5M parameters) on the challenging CA-LFW and CP-LFW benchmarks where the achieved accuracies by PocketNetS-128 are 95.48\% on CA-LFW and 89.63\% on CP-LFW in comparison to  95.15\% on CA-LFW and 88.55\% CP-LFW achieved by VarGFaceNet \cite{DBLP:conf/iccvw/YanZXZWS19}.

Our PocketNetM-128 (1.68M parameters) and PocketNetM-256 (1.75M parameters) outperformed all models proposed in the literature that have less than 2M  parameters. They also achieved competitive results to the models that have between 2 and 5M parameters, even outperforming them in many cases. For example, our PocketNetM-128 achieved SOTA accuracies on the challenging CA-LFW and CP-LFW among all models that have less than 5M of trainable parameters. 
On the large-scale evaluation benchmarks, IJB-B and IJB-C, our PocketNetM achieved competitive performance to many of the larger models. 
For example, on IJB-C,  our PocketNetM-128 (1.68M parameters) achieved verification performance of 92.63\% TAR at FAR 1e-6 and the best verification performance is 94.7\% achieved by MobileFaceNet \cite{martinez2021benchmarking} (2M parameters) and VarGFaceNet \cite{DBLP:conf/iccvw/YanZXZWS19} (5M parameters). 
On MegaFace and the refined version of MegaFace, our PocketNetM outperfomred all the models than have less than 2M of trainable parameters and they achieved a competitive results in term of identification and verification accuracies to the models that have between 2 and 5M parameters. For example, our PocketNetM-258 (1.75M parameters) outperformed MixFaceNet-S \cite{DBLP:conf/icb/BoutrosDFKK21} (3.07M parameters), ProxylessFaceNAS \cite{martinez2021benchmarking} (3.2M parameters) and MobileFaceNetV1 \cite{martinez2021benchmarking} (3.4M parameters) on MegaFace and MegaFace (R).

To visually illustrate the competence of our PocketNet, we plot the number of parameters vs. the achieved verification performance of our PocketNet and the recent compact models proposed in the literature (all numbers provided in Table \ref{tab:verification}). Figure \ref{fig:params} 
presents a trade-off between the number of parameters and the achieved verification performance.
Each of the presented solutions is marked with a point(x,y) in the plot, where x is the number of parameters in millions and y is the achieved verification performance. The model that tends to be placed on the top-left corner (small x and large y) of the plot has the best trade-off between the model compactness and the achieved verification performance. It can be observed, in Figure \ref{fig:params}, that our PocketNets are always in the top left corner in comparison to other methods, proving to achieve SOTA trade-off between model compactness and FR performance.
It must be noted that all the reported PocketNets in this section are trained with our proposed multi-step KD.

\vspace{-3mm}
\section{Conclusion}
\vspace{-1mm}
We present in this paper a family of extremely lightweight FR models, namely PocketNets. 
This is one of the first efforts proposing to utilize NAS to learn to design a compact yet accurate FR model. 
We additionally presented a novel training paradigm based on knowledge distillation, namely mulit-step KD, where the knowledge distillation is performed at multiple stages of the teacher training maturity. 
Extensive step-wise ablation studies proved the benefits of both, designing a face-specific architecture, as well as, the enhanced performance of the lightweight model when trained with the proposed multi-step KD.
Through extensive experimental evaluations on nine FR benchmarks, we demonstrated the high verification performance achieved by our compact PocketNet models and our proposed mulit-step KD.
Under the same level of model compactness, our PocketNets consistently scored SOTA performances in comparison to the compact models proposed in the literature. 

{\small
\bibliographystyle{ieee_fullname}
\bibliography{egpaper}
}

\end{document}